\documentclass[11pt, a4paper, onecolumn, copyright, gdm]{google}

\usepackage[authoryear, sort&compress, round]{natbib}
\uselogo{} 
\usepackage{subcaption}
\usepackage{latexsym}
\usepackage{tabularx}
\usepackage{multirow}
\newcommand{\PT}{\textsc{PT}}
\newcommand{\SFT}{\textsc{SFT}}
\newcommand{\corr}{\text{corr}}

\definecolor{DarkPink}{rgb}{0.5,0.0,0.18}
\definecolor{DarkGreen}{rgb}{0.1,0.5,0.1}
\definecolor{DarkRed}{rgb}{0.5,0.1,0.1}
\definecolor{DarkBlue}{rgb}{0.1,0.1,0.7}
\definecolor{DarkYellow}{rgb}{.79,.79,0}
\definecolor{1BOrange}{rgb}{0.90625,0.51171875,0.2265625}
\definecolor{240MGray}{rgb}{0.48046875,0.52734375,0.578125}
\definecolor{ForestGreen}{RGB}{34, 139, 34}
\definecolor{Commonsense}{RGB}{91, 155, 213}
\definecolor{Science}{RGB}{112, 173, 71}
\definecolor{NLI}{RGB}{158, 124, 195}
\definecolor{Paraphrase}{RGB}{237, 125, 49}
\definecolor{Reading}{RGB}{255, 217, 102}
\definecolor{Linguistic}{RGB}{123, 135, 148}
\definecolor{Semantic}{RGB}{237, 125, 49}

\title{The \textit{Magic Correlations}: \\
Understanding Knowledge Transfer from Pretraining to Supervised Fine-Tuning}

\correspondingauthor{berivan@google.com}

\reportnumber{} 

\author[2,3]{Simin Fan}
\author[2]{Dimitris Paparas}
\author[2]{Natasha Noy}
\author[1]{Binbin Xiong}
\author[1]{Noveen Sachdeva}
\author[1]{Berivan Isik}

\affil[1]{\thepa{}{}}
\affil[2]{Google Research}
\affil[3]{EPFL}

\begin{abstract}
Understanding how language model capabilities transfer from pretraining to supervised fine-tuning (SFT) is fundamental to efficient model development and data curation. 
In this work, we investigate four core questions: 
\textbf{RQ1}. To what extent do accuracy and confidence rankings established during pretraining persist after SFT? 
\textbf{RQ2}. Which benchmarks serve as robust cross-stage predictors and which are unreliable?
\textbf{RQ3}. How do transfer dynamics shift with model scale? 
\textbf{RQ4}. How well does model confidence align with accuracy, as a measure of calibration quality? Does this alignment pattern transfer across training stages?\\
We address these questions through a suite of \textit{correlation protocols} applied to accuracy and confidence metrics across diverse data mixtures and model scales.
Our experiments reveal that transfer reliability varies dramatically across capability categories, benchmarks, and scales---with accuracy and confidence exhibiting distinct, sometimes opposing, scaling dynamics.
These findings shed light on the complex interplay between pretraining decisions and downstream outcomes, providing actionable guidance for benchmark selection, data curation, and efficient model development.
\end{abstract}

\begin{document}

\maketitle
\begin{figure}[ht!]
\centering
\includegraphics[width=0.85\columnwidth]{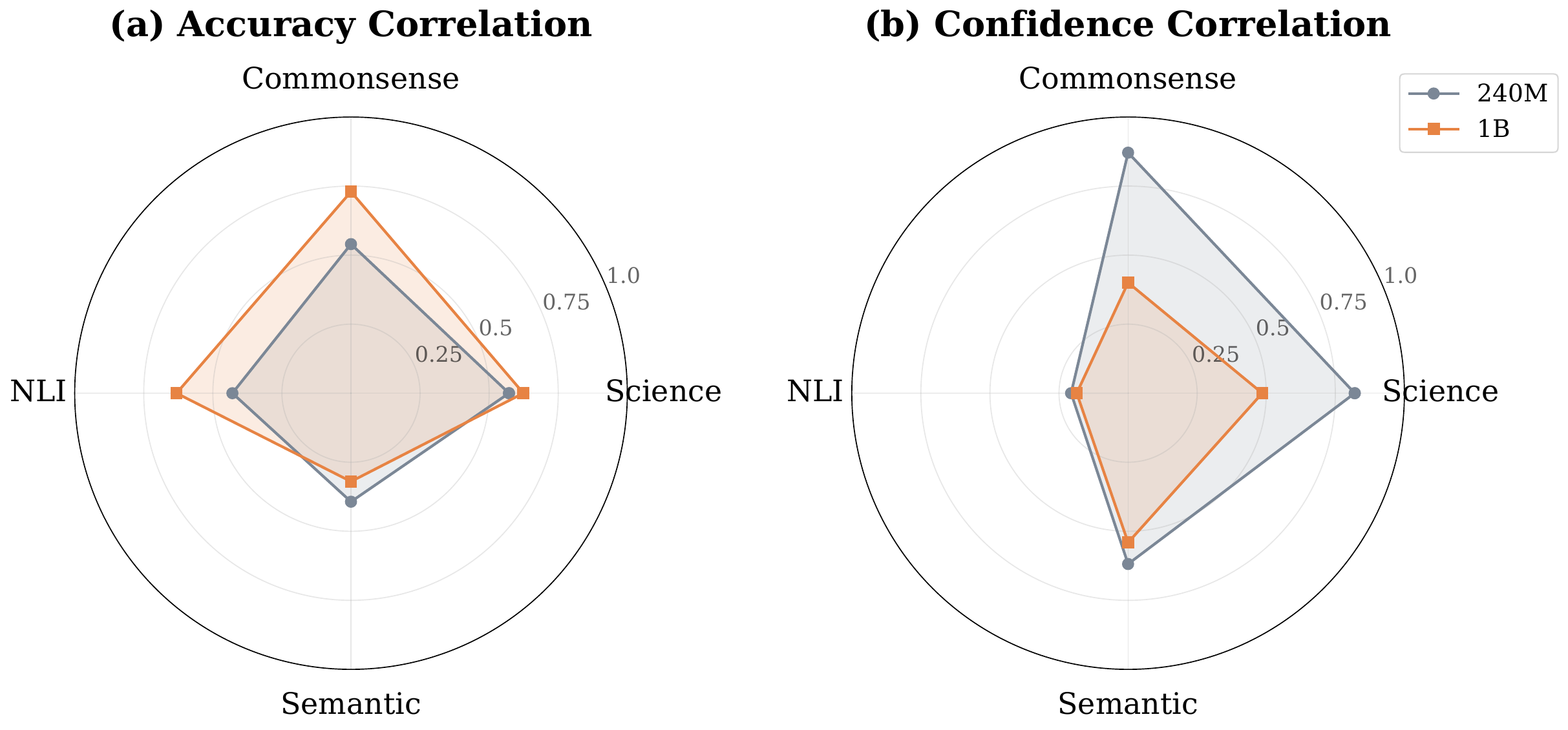}
\caption{\textbf{Cross-stage correlation by capability category.} \textit{(a) Accuracy correlation}: the 1B model generally
shows higher transferability; \textit{(b) Confidence correlation}: 240M maintains substantially higher correlation especially in Commonsense (\textcolor{240MGray}{0.87} vs.\ \textcolor{1BOrange}{0.40}) and Science (\textcolor{240MGray}{0.82} vs.\ \textcolor{1BOrange}{0.49}) domains.
This transferring pattern indicates that larger models undergo more confidence reorganization during 
SFT despite better accuracy preservation.}
\vspace{-2em}
\label{fig:radar_correlation}
\end{figure}
\section{Introduction}
Modern Large Language Model (LLM) training proceeds in stages: pretraining on massive text corpora~\citep{penedo2023refinedweb,weber2024redpajama,penedo2024fineweb,li2025datacomplmsearchgenerationtraining}, followed by supervised fine-tuning (SFT) on curated instruction data~\citep{ouyang2022training,ivison2023camelschangingclimateenhancing,chung2022scaling} and reinforcement learning~\citep{Guo_2025,wang2025reinforcementlearningreasoninglarge,cheng2025revisitingreinforcementlearningllm}. 
Critical decisions regarding data curation, mixture composition, and resource allocation are often made based on pretraining benchmarks alone, upon small-scale proxy models~\citep{kaplan2020scaling,hoffmann2022training,xie2023doremi,fan2024doge}.
This practice relies on the following fundamental assumption: pretraining performance reliably predicts post-SFT performance, and that internal model representations remain stable across training stages.

\noindent In this paper, we scrutinize this assumption through comprehensive empirical analysis of capability transfer from pretraining to SFT stage. We focus on four core questions:
\vspace{-0.5em}
\begin{enumerate}
    \item \textbf{Performance Stability}: To what extent do accuracy and confidence rankings established during pretraining persist after SFT?
    \item \textbf{Benchmark Reliability}: Which benchmarks serve as robust early-stage predictors, and which are unreliable?
    \item \textbf{Scale Dynamics}: How do transfer patterns shift with model scale?
    \item \textbf{Calibration Quality}: How well does model confidence align with accuracy? Does this alignment pattern transfer across training stages?
\end{enumerate}
\vspace{-0.5em}
To address these questions, we propose a suite of \textit{correlation protocols} and conduct systematic experiments on transformer models at two scales (240M and 1B parameters) across 9 diverse pretraining data mixtures. 
We evaluate on 20 benchmarks spanning four \textit{capability categories}, each corresponding to one particular skill and knowledge domain: \textbf{\textit{Commonsense Reasoning}} (\textcolor{Commonsense}{Commonsense}), \textbf{\textit{Scientific Reasoning}} (\textcolor{Science}{Science}), \textbf{\textit{Natural Language Inference}} (\textcolor{NLI}{NLI}), and \textbf{\textit{Semantic Understanding}} (\textcolor{Semantic}{Semantic}).

\noindent Our primary findings include:
\vspace{-0.5em}
\begin{itemize}
    \item \textbf{Inverse Scaling of Accuracy and Confidence Transfer}: 
    Larger models exhibit stronger cross-stage \emph{accuracy} correlation but weaker \emph{confidence} correlation, suggesting these metrics capture fundamentally different aspects of capability transfer. 
    \item \textbf{Task-Dependent Transfer Reliability}: 
    Transfer dynamics vary dramatically by capability domain. 
    \textit{Commonsense} and \textit{Science} categories show consistently high cross-stage correlation, while \textit{NLI} and \textit{Semantic Understanding} exhibit weaker patterns. 
    We identify specific benchmarks with particularly weak transfer, marking them as unreliable early-stage predictors.
    \item \textbf{Intra-Category Coherence Shifts with Scale}: 
    At smaller scales, benchmarks within the same capacity category often \emph{compete}---data mixtures that improve one benchmark degrade its semantic neighbors. 
    At larger scales, this competition gives way to \emph{synergy}, with positive intra-category correlations emerging.
    \item \textbf{Category-Dependent Calibration Quality}: 
    Performance-confidence alignment---the correlation between model confidence and accuracy---varies dramatically by capability domain, with \textit{Science} showing strong alignment while \textit{Commonsense} and \textit{Semantic} categories exhibit systematic miscalibration that persists through SFT.
    \item \textbf{Data Curation Trade-off between Accuracy and Calibration Quality}: 
    Strict educational-filtered pretraining data (Fineweb-edu) preserves or improve on scientific reasoning accuracy, while exhibits consistent performace-confidence alignment, revealing that strict educational filtering can preserves accuracy while degrades internal calibration patterns.
\end{itemize}
\vspace{-0.5em}
These findings provide actionable guidance for practitioners: \textit{benchmark selection} should account for category-specific transfer reliability and scale effects; \textit{pretraining data decisions} should consider cross-stage transferability and the divergent impacts on different capability domains.

\vspace{-0.5em}
\section{Related Work}
\label{sec:related}
\vspace{-0.5em}
Our work intersects several active research areas: benchmark design and valuations, transfer learning dynamics in LLMs, model calibration, and data curation.
\vspace{-0.5em}
\paragraph{Scaling Laws and Transfer Learning.}
The relationship between pretraining and downstream performance has been extensively studied through scaling laws~\citep{hernandez2021scaling,  chen2025scaling}. \citet{kaplan2020scaling} established power-law relationships governing how model performance improves with parameter count, dataset size, and compute budget. \citet{hoffmann2022training} refined it with the Chinchilla scaling laws, demonstrating that prior models were significantly undertrained relative to their parameter counts. More recently, \citet{hernandez2021scaling} extended scaling analysis to transfer learning settings, showing that SFT efficiency also follows predictable scaling relationships, with \citet{isik2025scaling, lourie2025scaling, schaeffer2025why} challenging these findings.
\vspace{-0.5em}
\paragraph{Benchmark Design and Contamination.}
The validity of evaluation benchmarks has received increasing scrutiny. \citet{sainz2023chatgpt} and \citet{deng2024investigating} documented widespread benchmark contamination in modern LLMs, where test data inadvertently appears in training corpora. \citet{rodriguez2021evaluation} analyzed benchmark difficulty and item response characteristics, while \citet{bowman2021fix} critiqued construct validity in NLU benchmarks. The HELM framework~\citep{liang2022holistic} introduced multi-dimensional evaluation spanning accuracy, calibration, robustness, and fairness. Our correlation-based analysis provides a complementary lens: rather than measuring absolute performance, we assess which benchmarks maintain \emph{stable rankings} across training stages---a property essential for reliable early-stage evaluation.

\vspace{-0.5em}
\paragraph{Model Calibration.}
Well-calibrated confidence estimates are crucial for deploying LLMs in high-stakes applications. \citet{guo2017calibration} demonstrated that modern neural networks are often poorly calibrated, and proposed temperature scaling as a post-hoc remedy. For language models specifically, \citet{kadavath2022language} showed that larger models exhibit improved calibration on factual questions, while \citet{desai2020calibration} found that pre-trained transformers require careful calibration for downstream tasks. \citet{tian2023just} and \citet{geng2024survey} provided comprehensive surveys of confidence estimation methods for LLMs.
Our cross-stage confidence correlation analysis extends this literature by examining whether calibration patterns \emph{persist} through SFT---a question with direct implications for practitioners who must decide when to re-calibrate models.
\vspace{-0.5em}
\paragraph{Data Mixture and Curation Effects.}
Understanding how pretraining data composition affects downstream capabilities is central to efficient LLM development. \citet{penedo2023refinedweb} demonstrated that carefully filtered web data can match curated corpora, while \citet{lozhkov2024fineweb-edu} showed that educational content filtering improves certain reasoning benchmarks. \citet{li2025datacomplmsearchgenerationtraining} introduced systematic data curation pipelines with measurable downstream effects. Data attribution methods~\citep{grosse2023studying, park2023trak} enable fine-grained analysis of which training examples contribute to specific capabilities. Our work takes a complementary aggregate approach: rather than attributing individual samples, we characterize how broad data mixture choices (web source, code proportion) affect the \emph{transferability} of capabilities across training stages.


\section{Experimental Protocols}
\label{sec:method}
\subsection{Training Pipeline}
We investigate the capability transfer dynamics between two critical phases of LLM development: \textit{Pretraining} (PT), where the model acquires foundational knowledge from massive textual corpora, and \textit{Supervised Fine-tuning} (SFT), where the model is adapted via instruction-following data for a more careful calibration.

\vspace{-0.5em}
\paragraph{Pretraining.} 
We train a suite of decoder-only transformer models at two scales ($240$M and $1$B parameters) on 9 distinct data mixtures (\autoref{tab:data_mixtures}). 
To systematically disentangle the effects of data source versus data distribution, we construct these mixtures by crossing three \textit{web data sources} and two \textit{code data sources}, with three \textit{mixing proportions}. 

\noindent Specifically, we utilize: (1) \textit{General Web Data} sourced from RefinedWeb~\citep{penedo2023refinedweb}, FineWeb-Edu~\citep{lozhkov2024fineweb-edu}, or DCLM~\citep{li2025datacomplmsearchgenerationtraining}; (2) \textit{Code Data} from StarCoder~\citep{li2023starcoder} or The Stack v2~\citep{lozhkov2024starcoder}; and (3) \textit{Curated Knowledge} from RedPajama-v2~\citep{weber2024redpajama} (including Wikipedia, ArXiv, Github and StackExchange).
The mixing proportions vary the ratio of web data ($25\%$, $45\%$, $65\%$) relative to code and curated sources, allowing us to comprehensively study the impact of data mixture composition on downstream knowledge transfer. Complete mixture specifications are provided in Appendix~\ref{app:data_mixtures}.

\vspace{-0.5em}
\paragraph{Supervised Fine-tuning (SFT).} 
Following the PT stage, we fine-tune the checkpoints pretrained from various pretraining data mixture on \texttt{Tulu-v2-mix}~\citep{ivison2023camelschangingclimateenhancing}. Models are pretrained for 12B (240M) and 52B (1B) tokens, followed by 5 epochs of SFT with cosine learning rate scheduling. Hyperparameter configurations are detailed in \autoref{app:setup}.

\subsection{Evaluation Benchmarks}
We evaluate the performance of language models across 20 benchmarks organized into four \textit{capability categories}, corresponding to various knowledge domains:
\vspace{-0.5em}
\begin{enumerate}
    \item \textbf{\textit{Commonsense Reasoning}} (\textcolor{Commonsense}{Commonsense}): CommonsenseQA~\citep{talmor2019commonsenseqaquestionansweringchallenge}, WinoGrande~\citep{sakaguchi2020winogrande}, HellaSwag~\citep{zellers2019hellaswag}, PIQA~\citep{bisk2020piqa}, SIQA~\citep{sap2019socialiqacommonsensereasoningsocial}, COPA~\citep{wang2019superglue}, BoolQ~\citep{wang2019superglue}---\textit{tasks requiring physical world knowledge and reasoning skill on social relationship};
    \item \textbf{\textit{Scienctific Reasoning}} (\textcolor{Science}{Science}): ARC-Challenge and ARC-Easy~\citep{clark2018arc}, SciQ~\citep{welbl2017crowdsourcingmultiplechoicescience}, OpenBookQA~\citep{mihaylov2018openbookqa}---\textit{tasks requiring factual and scientific knowledge};
    \item \textbf{\textit{Natural Language Inference}} (\textcolor{NLI}{NLI}): MNLI~\citep{williams2018mnli}, QNLI~\citep{wang2019gluemultitaskbenchmarkanalysis}, RTE~\citep{wang2019superglue}, CB~\citep{wang2019superglue}---\textit{tasks requiring inference about entailment relationships};
    \item \textbf{\textit{Semantic Understanding}} (\textcolor{Semantic}{Semantic}): QQP~\citep{sharma2019qqp}, MRPC~\citep{dolan2005mrpc}, WiC~\citep{pilehvar2019wic}, WSC~\citep{levesque2012wsc}, MultiRC~\citep{khashabi2018multirc}---\textit{tasks requiring understanding of semantic equivalence, coreference, and textual relationships}.
\end{enumerate}
\vspace{-0.5em}
We provide full benchmark details in Appendix~\ref{app:benchmarks}.
For each benchmark, we report two primary metrics: \textbf{\textit{accuracy}}, defined as the proportion of correct predictions; and \textbf{\textit{confidence}}, defined as the probability assigned to the selected answer averaged across the evaluation set.
\begin{figure*}[t!]
\centering

\includegraphics[width=\textwidth]{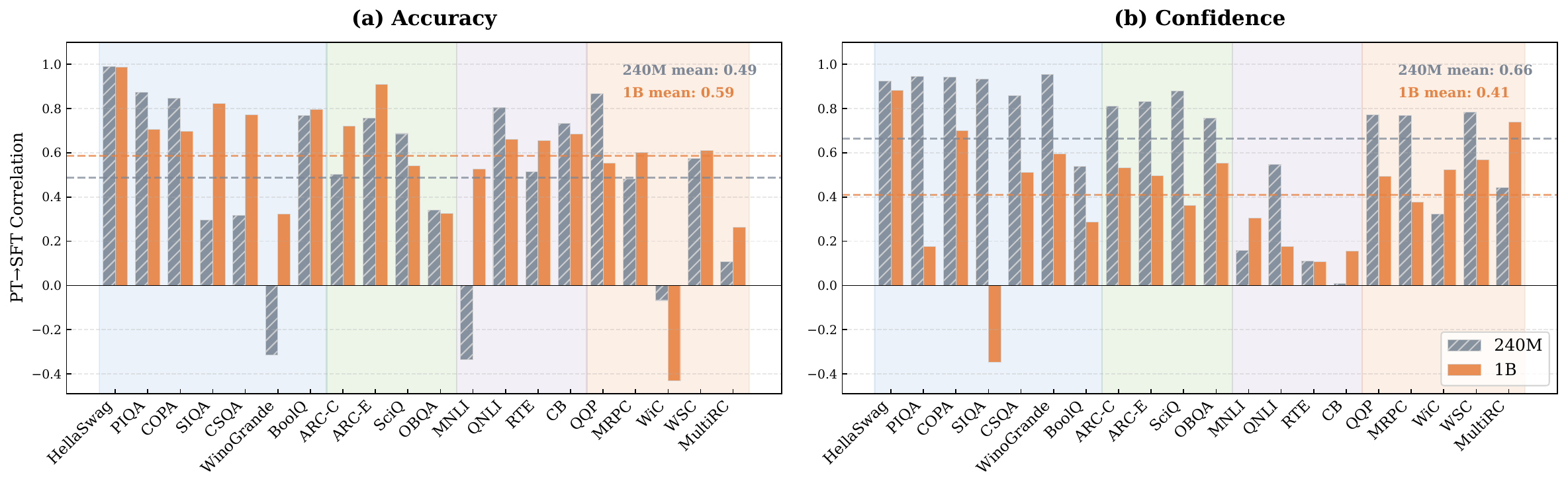}
\caption{\textbf{Cross-stage Correlation across various benchmarks.} Each bar shows the Pearson correlation between PT and SFT performance on the certain benchmark across data mixtures. (a)~\textit{Accuracy Correlation}: the \textcolor{1BOrange}{1B model} achieves higher transferrability than \textcolor{240MGray}{240M} (in average $\bar{r}$=\textcolor{1BOrange}{$\small0.59$} v.s. \textcolor{240MGray}{$\small0.49$}). 
(b)~\textit{Confidence Correlation}: the pattern \emph{reverses}---\textcolor{240MGray}{240M} achieves substantially stronger transfer than \textcolor{1BOrange}{1B model} ($\bar{r}$=\textcolor{1BOrange}{$\small0.41$} v.s. \textcolor{240MGray}{$\small0.66$}). Background colors indicate capability categories (\textcolor{Commonsense}{Commonsense}, \textcolor{Science}{Science}, \textcolor{NLI}{NLI}, \textcolor{Semantic}{Semantic}).}
\vspace{-1em}
\label{fig:acc_conf_transfer}
\end{figure*}

\subsection{The Lens of Correlations}
\label{sec:correlation_lens}
To quantify the preservation and reorganization of capabilities across training stages, we analyze model behavior through five complementary correlation protocols, along three dimensions.

\subsubsection{Cross-Stage Correlation}
We propose the following correlation protocols to measure \textit{how the accuracy and confidence calibration patterns established during pretraining persist through SFT}.

\vspace{-0.5em}
\paragraph{Cross-Stage Accuracy Correlation ($r_{\text{acc}}^{\text{stage}}$).}
For a given benchmark, we compute the Pearson correlation between pretraining and SFT accuracy across data mixtures:
\begin{equation}
    r_{\text{acc}}^{\text{stage}} = \corr\left(\mathbf{a}^{\PT}, \mathbf{a}^{\SFT}\right)
\end{equation}
where $\mathbf{a}^{\PT}, \mathbf{a}^{\SFT} \in \mathbb{R}^{M}$ are pretraining and SFT accuracy scores collected across $M=9$ mixtures.
High $r_{\text{acc}}^{\text{stage}}$ indicates the benchmark is a \textit{reliable early-stage proxy} from pretraining stage for downstream capability.

\vspace{-0.5em}
\paragraph{Cross-Stage Confidence Correlation ($r_{\text{conf}}^{\text{stage}}$).}
Analogously, for the confidence scores across various pretraining data mixture, we compute the cross-stage confidence correlation:
\begin{equation}
    r_{\text{conf}}^{\text{stage}} = \corr\left(\mathbf{c}^{\PT}, \mathbf{c}^{\SFT}\right)
\end{equation}
where $\mathbf{c}^{\text{PT}}$ and $\mathbf{c}^{\text{SFT}}$ represent the confidence scores across various data mixtures. A high $r_{\text{conf}}^{\text{stage}}$ suggests that the model's calibration ``fingerprint''---its level of uncertainty about specific inputs---derived from pretraining that persists despite the perturbations of SFT. 

\subsubsection{Intra-Category Correlation}\label{sec:intra-category-protocols}
Benchmarks are typically organized into semantic categories (e.g., commonsense reasoning, natural language inference) based on the assumption that tasks within a category tap similar underlying capabilities. But do benchmarks within the same category actually behave coherently when data mixtures change?
We investigate this question through three complementary protocols that capture different aspects of intra-category dynamics.

\vspace{-0.5em}
\paragraph{Pretraining Coherence ($r^{\PT}_{\text{intra}}$).}
For benchmarks $i, j$ in category $C$:
\begin{equation}
    r^{\PT}(i,j) = \corr\left(\mathbf{a}_i^{\PT}, \mathbf{a}_j^{\PT}\right), \quad i < j
\end{equation}
This metric measures how data mixtures that improve benchmark $i$ also improve benchmark $j$ \emph{during pretraining}. High positive values indicate \emph{pretraining-stage synergy}; negative values indicate \emph{competition} where optimizing for one benchmark degrades another.

\vspace{-0.5em}
\paragraph{SFT Coherence ($r^{\SFT}_{\text{intra}}$).}
Analogously for post-SFT:
\begin{equation}
    r^{\SFT}_{\text{intra}}(i,j) = \corr\left(\mathbf{a}_i^{\SFT}, \mathbf{a}_j^{\SFT}\right), \quad i < j
\end{equation}
Similarly, the within-SFT coherence assesses data mixtures that improve benchmark $i$ also improve benchmark $j$ \emph{after SFT}. Comparing $r^{\SFT}_{\text{intra}}$ with $r^{\PT}_{\text{intra}}$ reveals whether SFT \emph{introduces} or \emph{resolves} intra-category competition.

\vspace{-0.5em}
\paragraph{Cross-Stage Intra-Category Coherence ($r^{\text{cross}}_{\text{intra}}$).}
For transfer between \emph{different} benchmarks:
\begin{equation}
    r^{\text{cross}_{\text{intra}}}(i,j) = \corr\left(\mathbf{a}_i^{\PT}, \mathbf{a}_j^{\SFT}\right), \quad i \neq j
\end{equation}
The cross-stage transfer coherence answers whether the strong pretraining performance on benchmark $i$ can predict strong post-SFT performance on a \emph{different} benchmark $j$ within the same capability category. 
This measures \emph{cross-stage cross-benchmark transfer reliability}---whether capability improvements during pretraining generalize to related tasks after SFT.

For each protocol, we report the category-level average by averaging over all valid pairs within the same capability category:
\begin{equation}
    \bar{r}_{\text{intra}}(C) = \frac{2}{|C|(|C|-1)} \sum_{i < j,\; i,j \in C} r_{\text{intra}}(i,j)
\end{equation}

\subsubsection{Performance-confidence Alignment}
\label{sec:alignment_protocol}
Beyond transfer dynamics, we examine whether models develop well-calibrated representations through \textbf{\textit{performance-confidence alignment}}---the degree to which a model's confidence reflects its actual accuracy.
A model exhibits strong alignment when it is confident on correct predictions and uncertain on incorrect ones; poor alignment manifests as over-confidence on errors or under-confidence on correct answers.

\vspace{-0.5em}
\paragraph{Accuracy-Confidence Correlation ($r_{\text{align}}$).}
To quantify alignment quality, we correlate accuracy and confidence across benchmarks within a single model configuration:
\begin{equation}
    r_{\text{align}} = \corr\left(\mathbf{a}, \mathbf{c}\right)
\end{equation}
where $\mathbf{a}$ and $\mathbf{c}$ denote accuracy and confidence vectors across the benchmark suite.
This metric serves as a \textit{proxy for global calibration quality}: High $r_{\text{align}}$ indicates the model is confident when correct and uncertain when incorrect, indicating a well-calibrated model.
We use this metric to identify which pretraining data sources yield representations best aligned with target task demands.

\section{Results \& Findings}
\label{sec:results}
We present empirical observations centered around the four core research questions: 
\textbf{\textit{Cross-stage transfer}} (\S\ref{sec:cross_stage}), examining whether pretraining metrics persist after SFT;
\textbf{\textit{Benchmark reliability}} (\S\ref{sec:benchmark_reliability}), identifying which benchmarks serve as robust predictors across various capacity categories;
\textbf{\textit{Scaling dynamics}} (\S\ref{sec:scaling}), analyzing how transfer patterns shift with model scale; and
\textbf{\textit{Performance-confidence alignment}} (\S\ref{sec:alignment}), assessing calibration quality through accuracy-confidence correlation and its cross-stage transferability.

\subsection{Cross-Stage Transfer (RQ1)}
\label{sec:cross_stage}
We first examine to what extent the task performance (\textit{accuracy}) and calibration pattern (\textit{confidence}) established during pretraining persist after SFT.
We present the full results on cross-stage statistics and analysis in Appendix~\ref{app:cross_stage_results}.

\vspace{-0.5em}
\paragraph{Finding 1: Transfer reliability is strongly category-dependent.}
Figure~\ref{fig:radar_correlation}(a) reveals that capability categories exhibit markedly different transfer profiles.
Specifically, \textcolor{Science}{\textit{Science}} and \textcolor{Commonsense}{\textit{Commonsense}} benchmarks show consistently high cross-stage accuracy correlation across both model scales ($\bar{r} > 0.5$), indicating that pretraining performance on these domains reliably predicts post-SFT outcomes.
In contrast, \textcolor{Semantic}{\textit{Semantic}} tasks demonstrate weaker transferability, suggesting that SFT fundamentally reorganizes how models approach these linguistically nuanced tasks rather than building upon pretraining representations.

\begin{figure}[ht!]
\centering
\includegraphics[width=0.7\columnwidth]{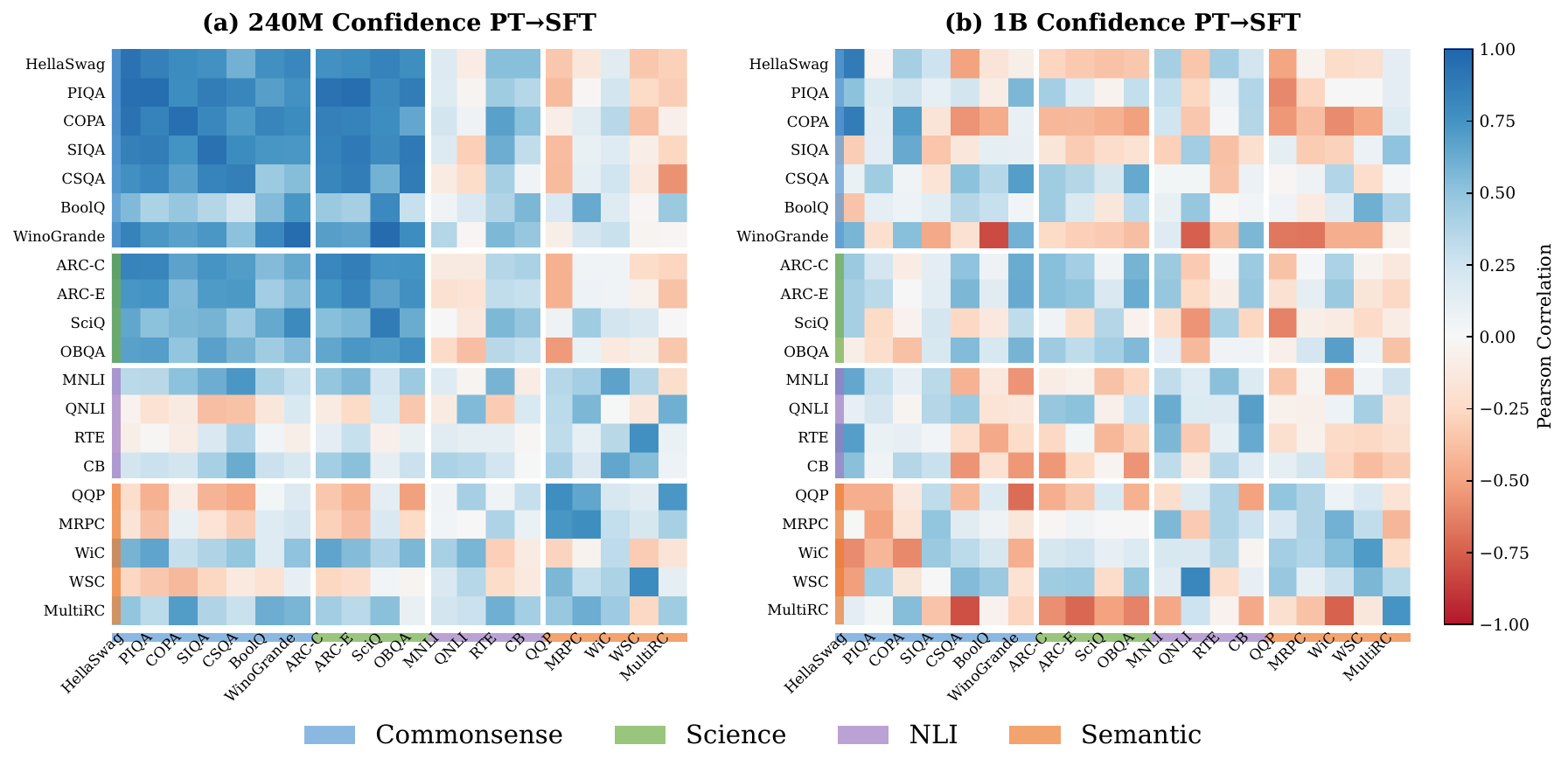}
\caption{\textbf{Cross-stage confidence correlation (PT$\to$SFT).} Each cell shows the Pearson correlation between benchmark $i$'s PT confidence and benchmark $j$'s SFT confidence across data mixtures; the diagonal represents the benchmark-wise transfer pattern.
\textbf{Left}: At 240M, the \textcolor{Commonsense}{\textit{Commonsense}}--\textcolor{Science}{\textit{Science}} block shows high \textcolor{DarkBlue}{\textbf{\textit{positive}}} correlations;
\textbf{Right}: At 1B, greater heterogeneity emerges with \textcolor{DarkRed}{\textbf{\textit{negative}}} correlations.}
\label{fig:conf_heatmap_cross}
\end{figure}

\vspace{-0.5em}
\paragraph{Finding 2: Confidence patterns persist more strongly than accuracy for reasoning tasks.}
Beyond the commonly-adopted accuracy metric, we examine whether confidence calibration established during pretraining persists after SFT.
Analyzing the diagonal of the PT$\to$SFT confidence correlation matrix (Figure~\ref{fig:conf_heatmap_cross}), we find remarkably strong persistence, especially at 240M model scale: mean benchmark-wise confidence correlation is $\bar{r} = 0.68$, with all evaluated benchmarks showing positive cross-stage correlation and 79\% exceeding $r > 0.5$.

This persistence is strongly category-dependent.
\textcolor{Commonsense}{\textit{Commonsense}} and \textcolor{Science}{\textit{Science}} benchmarks exhibit strong cross-stage confidence transfer ($\bar{r}$ = \textcolor{Commonsense}{$0.87$}, \textcolor{Science}{$0.82$}), indicating that a model's confidence profile on physical, social and scientific reasoning is largely determined during the pretraining stage.
In contrast, \textcolor{NLI}{\textit{NLI}} tasks show weak confidence persistence ($\bar{r} = 0.21$), suggesting that SFT substantially re-calibrates uncertainty on natural language inference tasks.

The practical implication is significant: for Commonsense and Science benchmarks, \textit{confidence-based model selection during pretraining remains a valid proxy for post-SFT calibration quality}, enabling early identification of well-calibrated models without expensive SFT iterations.

\begin{figure}[ht!]
\centering
\begin{subfigure}[b]{0.47\columnwidth}
    \includegraphics[width=\textwidth]{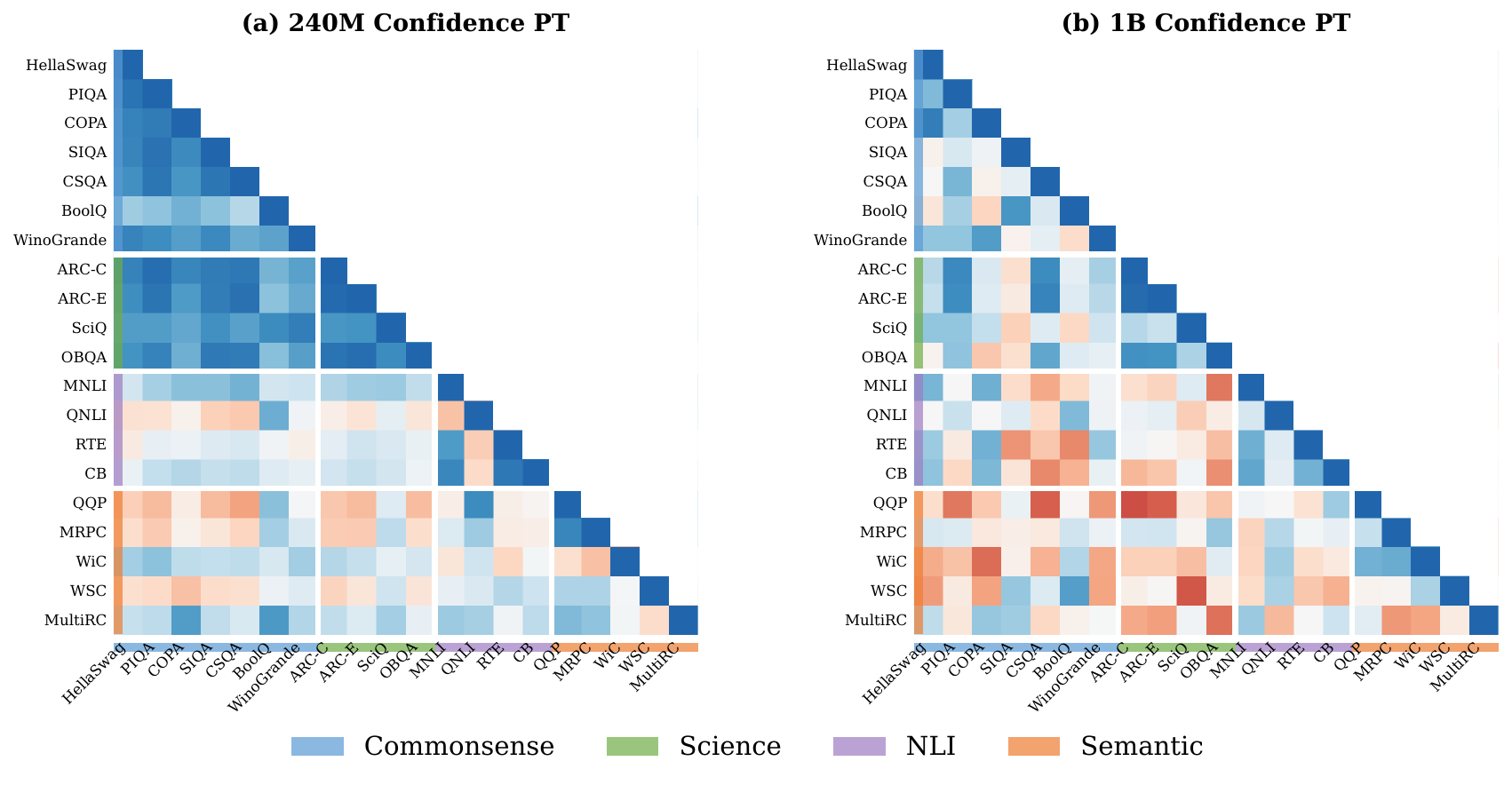}
    \caption{PT confidence correlation}
    \label{fig:conf_heatmap_pt}
\end{subfigure}
\hfill
\begin{subfigure}[b]{0.50\columnwidth}
    \includegraphics[width=\textwidth]{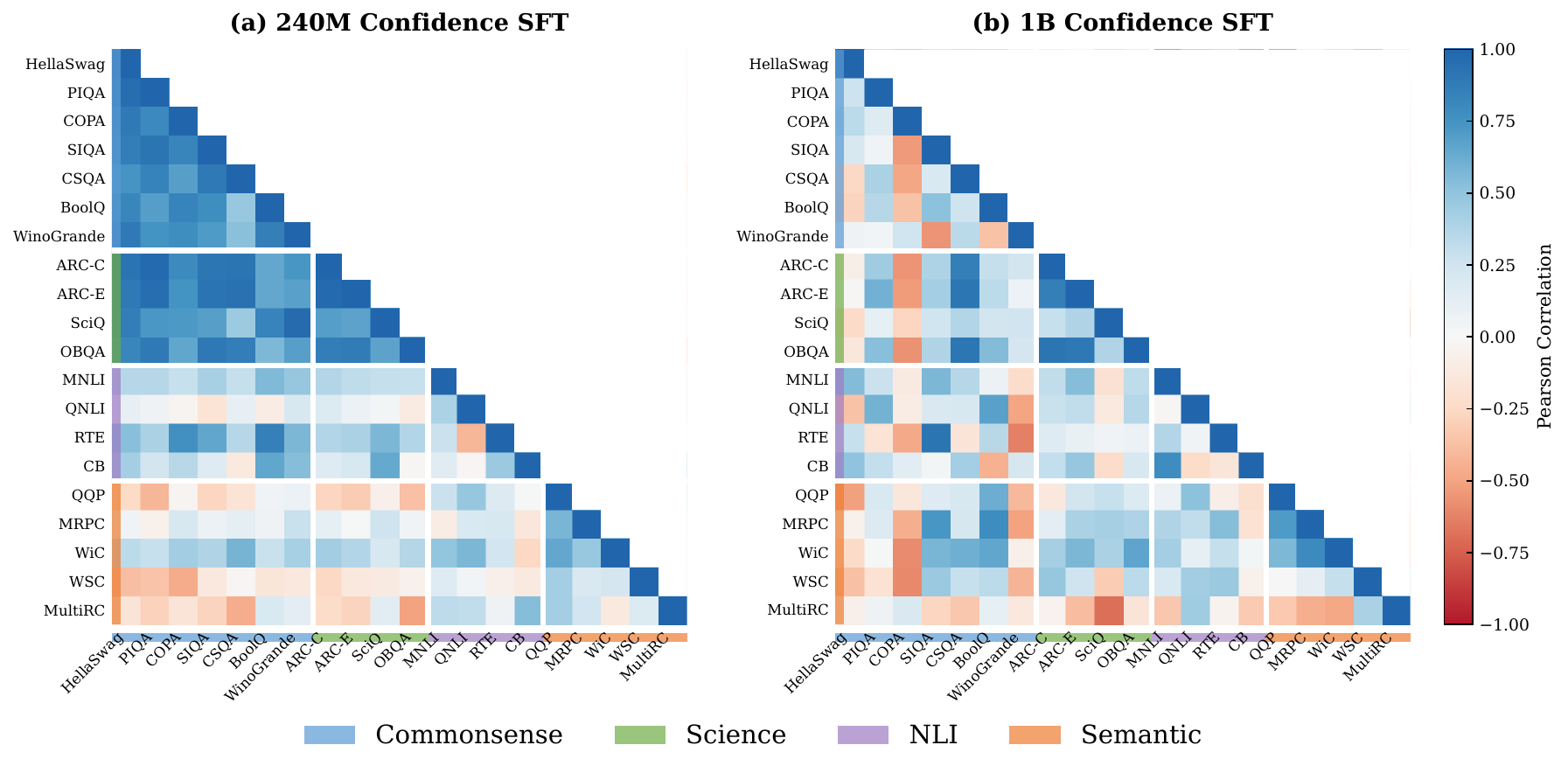}
    \caption{SFT confidence correlation}
    \label{fig:conf_heatmap_sft}
\end{subfigure}
\caption{\textbf{Within-stage confidence correlation comparison.}
(a) PT-PT and (b) SFT-SFT cross-benchmark confidence correlations at 240M and 1B scale.
The \textcolor{Commonsense}{\textit{Commonsense}}--\textcolor{Science}{\textit{Science}} block structure is nearly identical across stages, demonstrating that confidence correlation patterns established during pretraining persist through SFT.}
\label{fig:conf_heatmap_comparison}
\end{figure}

\vspace{-0.5em}
\paragraph{Finding 3: Cross-benchmark confidence structure persists across training stages.}
A stronger form of persistence emerges when comparing the \emph{structure} of confidence correlations across stages.
The PT-PT heatmap (Figure~\ref{fig:conf_heatmap_pt}) and SFT-SFT heatmap (Figure~\ref{fig:conf_heatmap_sft}) demonstrate the cross-benchmark predictability of confidence scores within the same training stage. 
According to Figure~\ref{fig:conf_heatmap_pt} and \ref{fig:conf_heatmap_sft}, at 240M exhibit striking structural similarity: these two within-stage correlation matrices are highly correlated with $r = 0.73, p<10^{-32}$, indicating that \textit{benchmark pairs that co-vary in confidence during pretraining continue to co-vary similarly after SFT}.

This cross-benchmark cross-stage consistency is particularly strong for \textcolor{Commonsense}{\textit{Commonsense}} and \textcolor{Science}{\textit{Science}} categories.
The \textit{Commonsense}--\textit{Science} ``confidence block'' not only persists but slightly \emph{strengthens} through SFT, suggesting that SFT reinforces rather than disrupts the shared calibration structure between these reasoning categories.

\subsection{Benchmark Reliability (RQ2)}
\label{sec:benchmark_reliability}
We further conduct benchmark-wise analysis to investigate which benchmarks serve as reliable predictors of post-SFT performance and which should be treated with caution during early-stage evaluation. 
Detailed statistics are provided in Appendix~\ref{app:per_benchmark}.

\vspace{-0.5em}
\paragraph{Finding 4: Certain benchmarks are unreliable cross-stage predictors.}
Figure~\ref{fig:acc_conf_transfer}(a) reveals that individual benchmarks vary dramatically in their predictive reliability.
WiC and MultiRC exhibit particularly weak or negative accuracy correlations ($r_{\text{acc}}^{\text{stage}} < 0.3$ across scales), indicating that strong pretraining performance on these benchmarks does not guarantee post-SFT success.
Similarly, WinoGrande and MNLI show negative transfer at 240M scale ($r = -0.31$ and $r = -0.34$ respectively).

These unreliable benchmarks share a common characteristic: they require nuanced linguistic understanding that appears to be substantially reorganized during instruction tuning.
\textit{Practitioners should exercise caution when using these benchmarks for early-stage model selection or data curation decisions.}

\vspace{-0.5em}
\paragraph{Finding 5: Intra-category competition undermines single-benchmark evaluation.}
Beyond individual benchmark reliability, we examine whether benchmarks within the same capacity category respond coherently to data mixture changes.
As shown in Figure~\ref{fig:intra_category}, at both model scales, \textcolor{Commonsense}{\textit{Commonsense}} and \textcolor{Semantic}{\textit{Semantic}} categories consistently exhibit negative intra-category coherence on accuracy.

This ``\textit{intra-category competition}'' implies that data mixtures optimizing one benchmark often degrade its semantic neighbors---a critical warning for practitioners who assume that improving HellaSwag indicates general improvements in Commonsense Reasoning.
Notably, \textcolor{Science}{\textit{Science}} demonstrate positive intra-category coherence on large-scale model (Figure~\ref{fig:intra_category} (b)), suggesting more unified underlying representations for scientific knowledge.

\vspace{-0.5em}
\paragraph{Finding 6: Certain benchmarks behave incoherently within their semantic categories.}
Beyond cross-stage reliability, we identify benchmarks that fail to co-vary with other benchmarks in the same capability category---acting as ``black sheep'' that respond differently to data mixture changes than semantically related tasks.
Within \textcolor{Commonsense}{\textit{Commonsense}} category, WinoGrande exhibits consistently negative correlations with other category members (mean pairwise $r = -0.18$ at 240M);
within \textcolor{Semantic}{\textit{Semantic}} group, WSC shows weak coherence with $r < 0.15$, indicating that coreference resolution may not share underlying representations with other linguistic understanding benchmarks.

These incoherent benchmarks pose a subtle risk: \textit{even when a benchmark shows reasonable cross-stage transfer individually, the low category coherence means that the task-specific improvements cannot generalize to semantically related tasks.}

\subsection{Scaling Dynamics (RQ3)}
\label{sec:scaling}
We now examine how transfer patterns shift as models scale from 240M to 1B, revealing systematic differences in accuracy and confidence patterns across scales.

\vspace{-0.5em}
\paragraph{Finding 7: Accuracy and confidence transfer exhibit inverse scaling dynamics.}
Figure~\ref{fig:acc_conf_transfer} reveals a striking dissociation between accuracy and confidence as models scale.
For \emph{accuracy}, the 1B model achieves stronger cross-stage transfer than 240M ($\bar{r}_{\text{acc}}$ = \textcolor{1BOrange}{$0.60$} vs.\ \textcolor{240MGray}{$0.51$}), consistent with the intuition that larger models learn more transferable representations.
However, for \emph{confidence}, the pattern \emph{reverses}: 240M maintains substantially higher cross-stage correlation ($\bar{r}_{\text{conf}}$ = \textcolor{240MGray}{$0.68$} vs.\ \textcolor{1BOrange}{$0.39$}).

This inverse scaling dynamic suggests that larger models undergo more extensive confidence reorganization during SFT, developing task-specific calibration profiles rather than retaining the monolithic uncertainty patterns observed on small-scale models.
The implication for practitioners is significant: \textit{as models scale, confidence calibration from pretraining becomes less predictive of post-SFT behavior, warranting explicit calibration techniques for SFT.}

\begin{figure}[ht!]
\centering
\includegraphics[width=0.7\columnwidth]{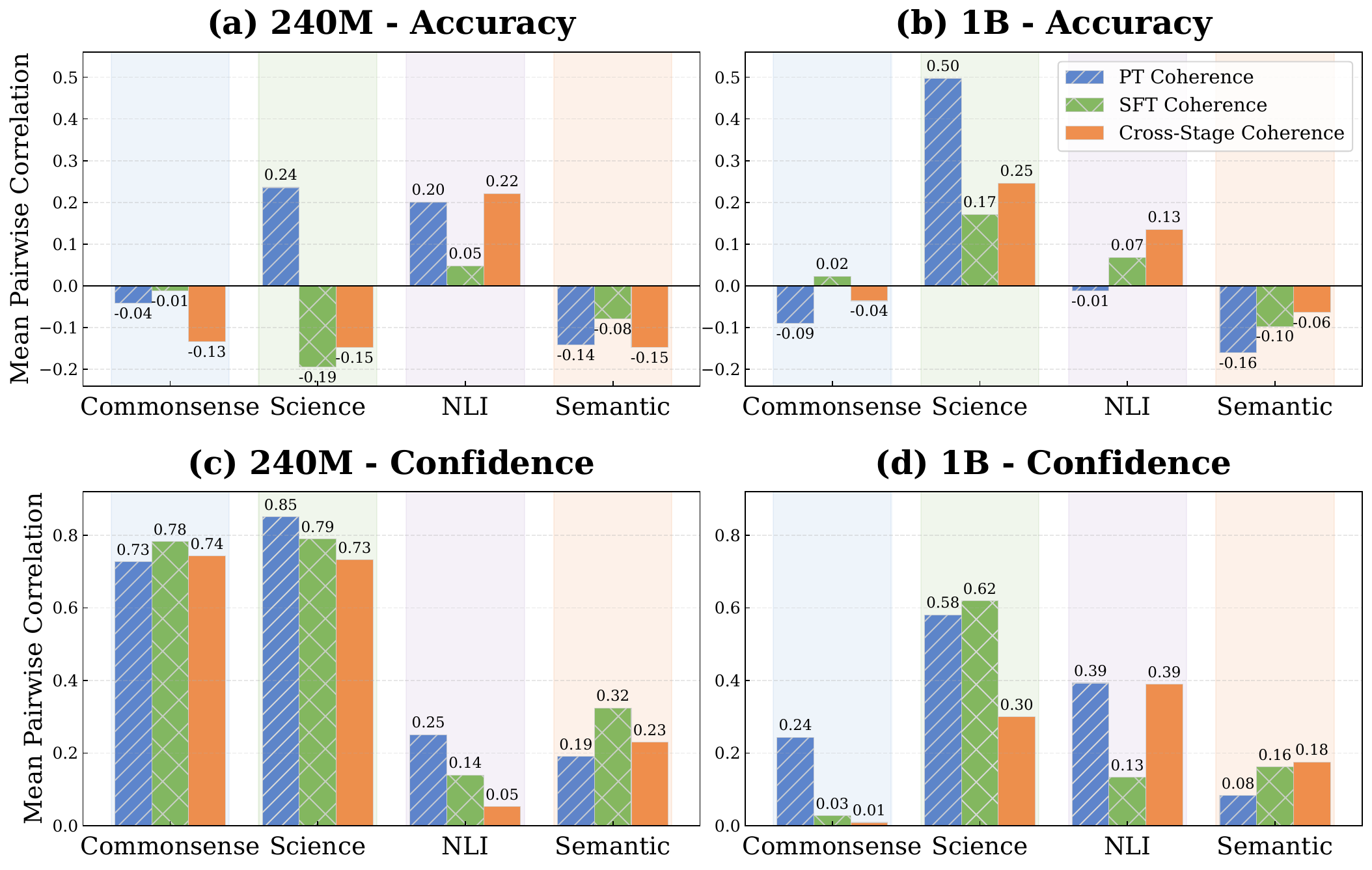}
\caption{\textbf{Intra-category coherence across three correlation protocols.} 
\textbf{Top}: the coherence scores on \textit{accuracy}, where \textcolor{Science}{Science} shows PT$\to$SFT degradation; \textcolor{NLI}{NLI} preserves cross-stage coherence despite a drop in SFT coherence.
\textbf{Bottom}: the coherence scores on \textit{confidence}, where 240M model maintains high coherence while 1B shows substantial degradation.}
\vspace{-1.5em}
\label{fig:intra_category}
\end{figure}

\vspace{-0.5em}
\paragraph{Finding 8: Scaling induces a transition from intra-category competition to synergy in accuracy.}
Figure~\ref{fig:intra_category} reveals that model scale fundamentally alters intra-category dynamics for accuracy.
For both \textit{SFT coherence} and \textit{cross-stage coherence}, most capability categories shift toward less negative or positive coherence at 1B scale.
The \textcolor{Science}{\textit{Science}} category exhibits particularly significant improvements across all three protocols with 240M$\to$1B scaling (PT: $0.24 \to 0.50$; SFT: $-0.19 \to 0.17$; Cross-stage: $-0.15 \to 0.25$).
This suggests that larger models develop more unified reasoning representations where knowledge transfers constructively across related benchmarks.
It implies that \textit{at smaller scales, optimizing for a single benchmark per category may harm related tasks; while at larger scales, the generalization within the same capacity category becomes more viable.}

\vspace{-0.5em}
\paragraph{Finding 9: Confidence coherence degrades with scale while remaining positive.}
For confidence calibration, we observe the opposite scaling trend (Figure~\ref{fig:intra_category}).
At 240M, \textcolor{Commonsense}{\textit{Commonsense}} and \textcolor{Science}{\textit{Science}} exhibit strong positive coherence ($\bar{r} = 0.73$ and $0.85$ respectively); while at 1B, this coherence diminishes substantially ($\bar{r} = 0.24$ and $0.58$).

Rather than indicating performance degradation, this drop reflects the development of more \emph{task-specific uncertainty estimates} on larger language models---when scaling-up, the model learns to calibrate confidence differently for distinct tasks or benchmarks, instead of adopting a universal uniform calibration pattern of smaller models.

\vspace{-0.5em}
\paragraph{Finding 10: NLI exhibits anomalous scaling patterns.}
The \textcolor{NLI}{NLI} category presents a distinct scaling patterns varying from other benchmarks described above.
For confidence coherence, NLI is the \emph{only} category where 1B shows higher coherence than 240M ---the inverse of the trend where scale typically reduces confidence coherence.

This twofold anomaly suggests that NLI tasks engage qualitatively different representational mechanisms. One hypothesis is that larger models develop more consistent inferential strategies for NLI specifically, improving calibration coherence even as other categories become more task-specifically calibrated.

\subsection{Calibration Quality (RQ4)}
\label{sec:alignment}
Finally, we examine \textit{performance-confidence alignment}---the correlation between model confidence and accuracy scores ($r_{align}$)---as a measure of \textit{calibration quality}. 
A well-calibrated model should be confident precisely when it is correct and uncertain when it is incorrect, yielding high $r_{align}$. 
We investigate how this alignment varies across capability categories, data mixtures, and training stages.

\begin{figure}[htbp!]
\centering
\includegraphics[width=0.7\columnwidth]{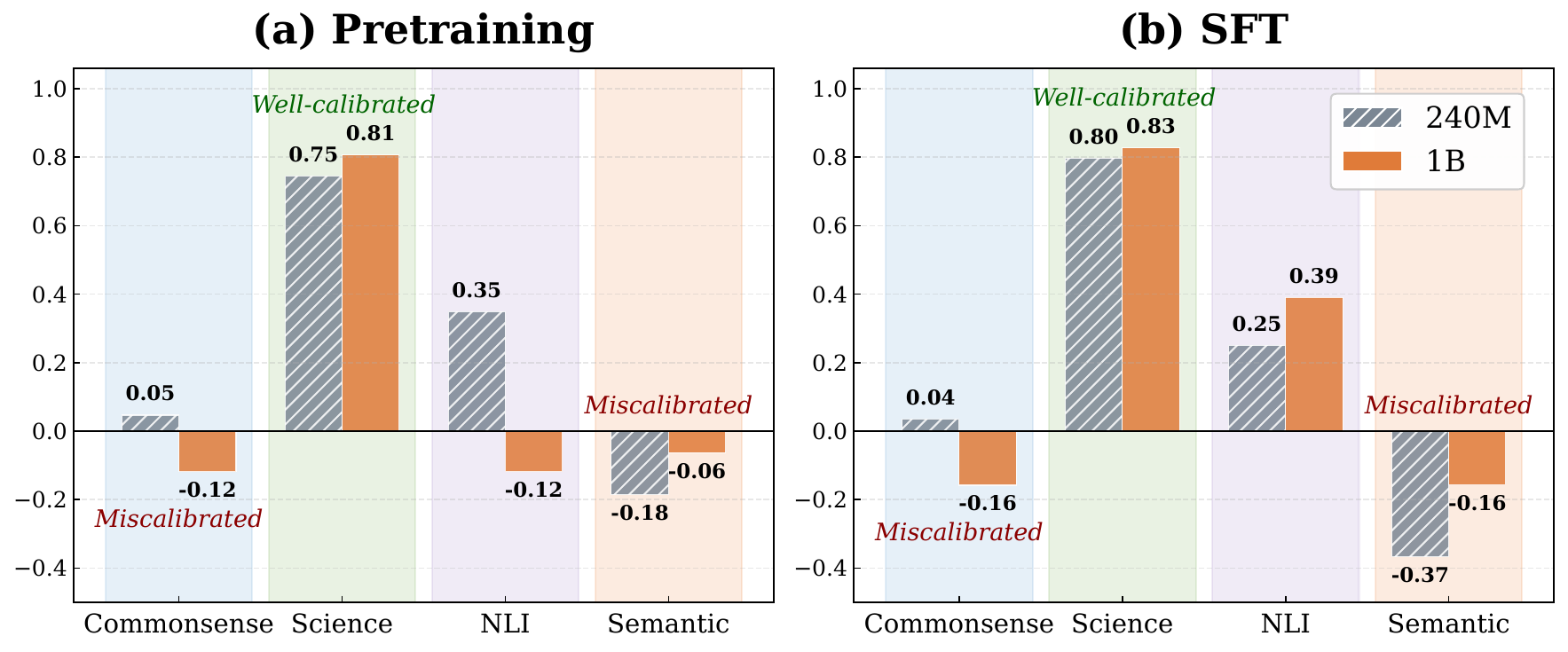}
\caption{\textbf{Performance-confidence alignment ($r_{\text{align}}$) varies by category.} 
\textcolor{Science}{Science} tasks show high positive alignment ($r_{\text{align}} \approx 0.8$), indicating well-calibrated confidence.
\textcolor{Commonsense}{Commonsense} and \textcolor{Semantic}{Semantic} tasks show negative alignment, indicating systematic overconfidence on incorrect predictions.}
\vspace{-1.5em}
\label{fig:knowledge_alignment}
\end{figure}

\vspace{-0.5em}
\paragraph{Finding 11: Performance-confidence alignment varies significantly across capability categories.}
Figure~\ref{fig:knowledge_alignment} reveals striking differences in how well models' confidence aligns with actual performance across capability domains.
\textcolor{Science}{\textit{Science}} tasks exhibit remarkably high alignment ($r_{\text{align}} \approx 0.8$), indicating that models are confident precisely when they perform well---the prior knowledge from pretraining aligns well with downstream task requirements.

In stark contrast, \textcolor{Commonsense}{\textit{Commonsense}} shows consistently weak or negative alignment ($r_{\text{align}} \approx -0.1$), meaning models tend to be \emph{more} confident on incorrect predictions.
\textcolor{Semantic}{\textit{Semantic}} tasks exhibit similar miscalibration ($r_{\text{align}} \approx -0.2$).
This systematic miscalibration suggests a fundamental mismatch between pretraining knowledge and the implicit reasoning required for commonsense understanding---the model's confidence reflects surface-level pattern matching rather than genuine comprehension.

\vspace{-0.5em}
\paragraph{Finding 12: Alignment patterns persist through SFT.}
Comparing the left and right panels of Figure~\ref{fig:knowledge_alignment}, the category-level alignment structure remains largely stable across training stages.
\textcolor{Science}{\textit{Science}} tasks maintain high $r_{\text{align}}$ values ($\sim$0.8) in both pretraining and post-SFT evaluation, while the miscalibration for \textcolor{Commonsense}{\textit{Commonsense}} and \textcolor{Semantic}{\textit{Semantic}} also persists.
This suggests that SFT does not fundamentally reorganize the model's confidence structure---the alignment ``fingerprint'' established during pretraining carries through to downstream performance.

Combined with our cross-stage analysis (\S\ref{sec:cross_stage}), this finding reinforces that \textit{pretraining data composition has lasting effects on model behavior that cannot be easily overwritten by instruction tuning}.
Practitioners seeking well-calibrated models should prioritize appropriate pretraining data curation over post-hoc calibration techniques.

\begin{figure}[ht!]
\centering
\includegraphics[width=0.7\columnwidth]{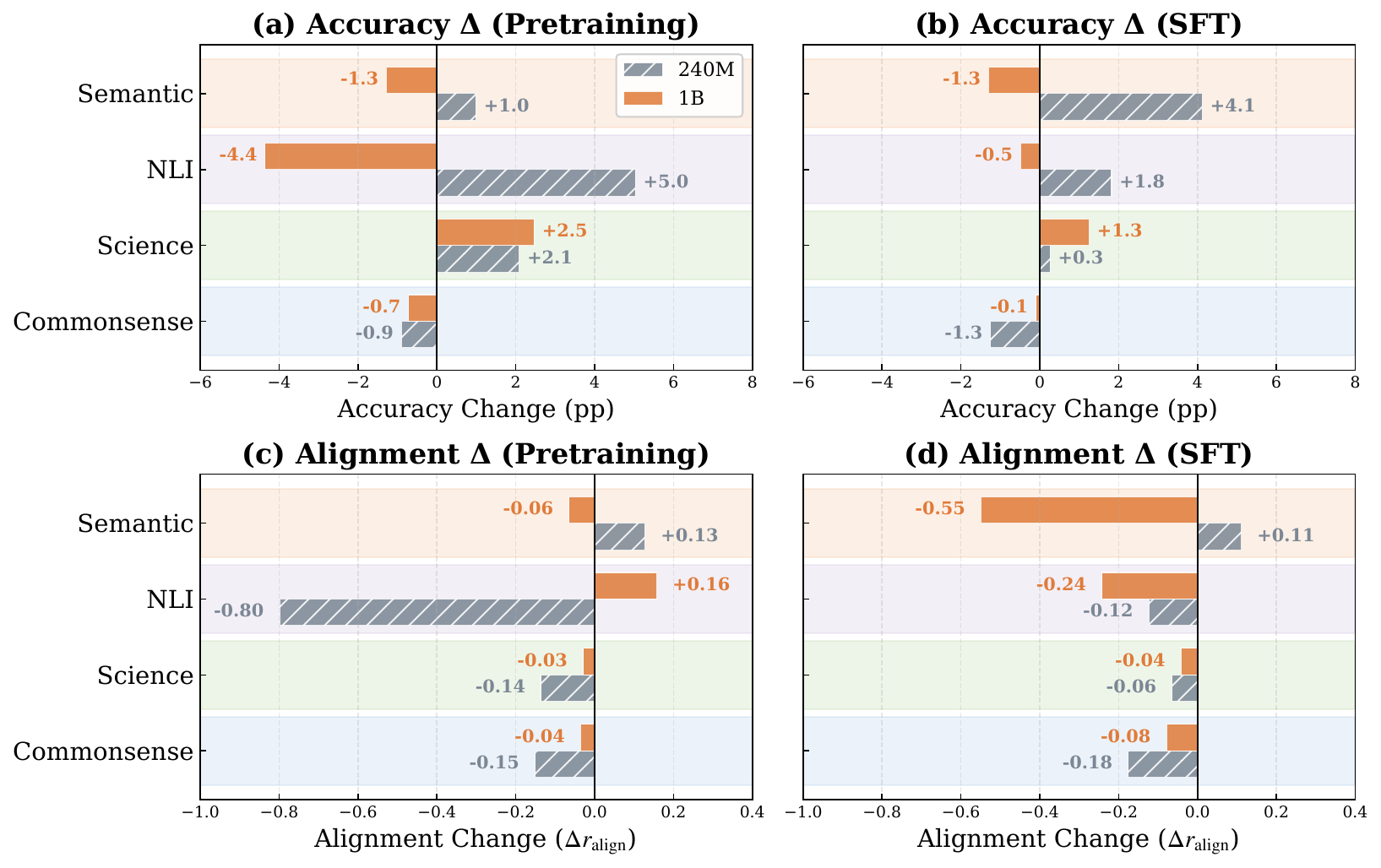}
\caption{\textbf{FineWeb-Edu vs.\ RefinedWeb: Category-level effects at PT and SFT stages.}
(a,b)~Accuracy changes; (c,d)~Alignment changes.
Left column: Pretraining; Right column: SFT.
The NLI accuracy reversal (240M: +5.0pp $\to$ 1B: $-4.4$pp) persists through SFT, though attenuated.
At 1B SFT, Semantic alignment shows severe degradation ($\Delta = -0.55$), suggesting that FineWeb-Edu's effects on linguistic tasks become increasingly problematic at larger scales and later training stages.}
\vspace{-1.em}
\label{fig:fineweb_edu_delta_full}
\end{figure}

\vspace{-0.5em}
\paragraph{Finding 13: Educational content filtering exhibits scale-dependent accuracy-calibration trade-offs.}
Figure~\ref{fig:fineweb_edu_delta_full} reveals that the effects of educational content filtering (FineWeb-Edu) vary dramatically with model scale.
At 240M, FineWeb-Edu \emph{improves} NLI accuracy by +5.0pp compared to RefinedWeb, yet simultaneously \emph{degrades} alignment from $r_{\text{align}} = 0.68$ to $r_{\text{align}} = -0.12$ ($\Delta = -0.80$).
Remarkably, at 1B scale, this pattern \emph{reverses}: FineWeb-Edu \emph{degrades} NLI accuracy by $-4.4$pp while slightly \emph{improving} alignment ($\Delta = +0.16$).
On \textcolor{Science}{\textit{Scientific Reasoning}} tasks, exhibits consistent benefits across scales: FineWeb-Edu improves accuracy at both 240M (+2.1pp) and 1B (+2.5pp), with slight alignment degradation.

It indicates that: \textit{data curation decisions optimized at small proxy scales may not transfer to larger models}; and \textit{strict educational filtering could preserves or improve on accuracy scores while undermining internal calibration pattern.}

\section{Practical Implications}
Our findings offer several actionable insights for practitioners developing and evaluating language models.

\paragraph{Benchmark selection for early-stage evaluation.}
Not all benchmarks are equally informative during pretraining evaluation.
According to the clear dichotomy we have observed: \textit{Commonsense Reasoning} benchmarks (e.g. HellaSwag, PIQA, COPA) and \textit{Scientific Reasoning} benchmarks exhibit high cross-stage correlation, making them reliable proxies for post-SFT performance; while \textit{Semantic Understanding} tasks (WiC, WSC, MultiRC) and \textit{NLI} benchmarks (MNLI) show weak or negative transfer, suggesting their pretraining performance provides limited signal about final model capabilities.

\noindent\textbf{Implication 1}: \textit{Practitioners should prioritize high-transfer benchmarks while avoiding unreliable ones when making early-stage decisions about data curation and resource allocation.}

\vspace{-0.5em}
\paragraph{Confidence as a Complementary Evaluation Signal.}
We demonstrate that confidence correlation provides information complementary to accuracy, with important scale-dependent dynamics.
While confidence patterns transfer reliably across stages at smaller scales (240M) reflecting a relatively consistent calibration profile that persists through SFT, confidence transferability degrades substantially when the model scaled up to 1B parameters.
This suggests that larger models develop more \textit{nuanced, task-specific calibration profiles} during SFT, flipping the monolithic uncertainty patterns of smaller models.

\noindent\textbf{Implication 2}: \textit{As models scale, practitioners cannot assume that well-calibrated pretraining confidence will carry over to post-SFT deployment. Task- and stage-specific calibration becomes increasingly critical at larger scales.}

\vspace{-0.5em}
\paragraph{Scale-Dependent Data Curation Effects.}
Our analysis reveals that aggressive educational content filtering (FineWeb-Edu) produces scale-dependent effects that challenge the common practice of using small proxy models for data curation decisions.
For NLI, FineWeb-Edu yields +5.0pp accuracy gains at 240M but $-4.4$pp \emph{degradation} at 1B---a complete reversal (Figure~\ref{fig:fineweb_edu_delta_full}).
Meanwhile, alignment effects also reverse: severe degradation at 240M ($\Delta = -0.80$) but slight improvement at 1B ($\Delta = +0.16$).

\noindent\textbf{Implication 3}: \textit{Practitioners should exercise caution when extrapolating data curation decisions from small-scale experiments to larger models.
For robust data curation, we recommend evaluating mixture effects at multiple scales before committing to production training.}

\section{Conclusion}
\label{sec:conclusion}
We systematically analyze capability transfer from pretraining to SFT using correlation-based protocols across diverse data mixtures, model scales, and a large variety of evaluation benchmarks.
Our findings provide practical guidance for early-stage benchmark selection and data mixture design, while highlighting the lasting impact of pretraining decisions on model behavior.
\vspace{-0.5em}
\paragraph{Limitations and Future Direction.}
Our study has several limitations that suggest directions for future work.
Because of resource constraints, our experiments are conducted at 240M and 1B parameter scales using a single SFT dataset (\texttt{Tulu-v2-mix}), which may not fully capture phenomena at larger scales or under different post-training regimes.
Additionally, while we evaluate 20 benchmarks across diverse capability categories, important domains such as long-context reasoning, code generation, and safety remain unexplored.
We provide a detailed discussion of limitations and future directions in Appendix~\ref{app:limitations}.

\newpage
\bibliography{main}
\newpage
\appendix
\appendix
\onecolumn
\clearpage
\newpage
\section{Limitations and Future Work}
\label{app:limitations}

\paragraph{Model Scale.}
Our experiments are conducted at 240M and 1B parameter scales, which may not fully capture phenomena that emerge at larger scales.
The inverse scaling trends we observe for accuracy versus confidence transfer are particularly intriguing and warrant investigation at 7B+ scales to determine whether these dynamics persist, amplify, or reverse.
We leave further scaling efforts beyond the \textit{small language model} (SLM) scope for future work.

\paragraph{SFT Data and Protocol.}
We use a single SFT dataset (\texttt{tulu-v2-mix}) and training protocol across all experiments.
Different instruction-tuning datasets or multi-stage post-training pipelines (e.g., incorporating RLHF or DPO) may exhibit different transfer dynamics.
Our findings are most directly applicable to standard SFT workflows and may not generalize to more complex post-training regimes.

\paragraph{Benchmark Coverage.}
While we evaluate on 22 benchmarks spanning 8 capability categories, our analysis necessarily omits many important capabilities, including long-context reasoning, multi-turn dialogue, code generation, and safety-related behaviors.
The transfer patterns we identify may not generalize to these domains, particularly those that are specifically targeted by instruction-tuning data.

\paragraph{Data Mixture Granularity.}
Our 9 data mixtures, while systematically varied, represent a sparse sampling of the possible mixture space.
Finer-grained variations in code proportion, web source combinations, or the inclusion of additional data types may reveal patterns not captured in our analysis.
Applying our proposed correlation protocols to study the impact of \textbf{\textit{knowledge-enriched content}} such as textbooks~\citep{gunasekar2023textbooksneed} or \textbf{\textit{synthetic data}}~\citep{kang2025demystifyingsyntheticdatallm,chen2024diversitysyntheticdataimpact,datologyai2025beyondweblessonsscalingsynthetic} across training stages is a promising direction for future work.

\clearpage
\newpage
\section{Experimental Setup}
\label{app:setup}

\subsection{Model Architecture}
\label{app:architecture}

We train decoder-only transformer models at two parameter scales.
Table~\ref{tab:model_config} summarizes the architectural configurations.

\begin{table}[ht!]
\centering
\caption{Model architecture configurations.}
\label{tab:model_config}
\small
\begin{tabular}{lcc}
\toprule
\textbf{Hyperparameter} & \textbf{240M} & \textbf{1B} \\
\midrule
Model dimension & 768 & 1680 \\
Per head dimension & 256 &  256\\
Number of heads & 8 & 8 \\
Number of layers & 8 & 12 \\
Expand factor & 8 & 8 \\
Context length & 4096 & 4096 \\
Vocabulary size & 100,864 & 100,864 \\
\midrule
Total parameters & 240M & 1.0B \\
\bottomrule
\end{tabular}
\end{table}

\subsection{Training Hyperparameters}
\label{app:hyperparameters}

\paragraph{Pretraining.}
Table~\ref{tab:pt_hyperparams} details the pretraining hyperparameters for both model scales.
We use the AdamW optimizer with a cosine learning rate schedule and linear warmup.

\paragraph{Supervised Fine-tuning.}
Following pretraining, we fine-tune each checkpoint on \texttt{tulu-v2-mix}~\citep{ivison2023camelschangingclimateenhancing}, a curated collection of instruction-following datasets including FLAN-v2, Open Assistant, ShareGPT, and others.
Table~\ref{tab:sft_hyperparams} summarizes SFT hyperparameters.

\begin{table*}[ht!]
\centering
\caption{Training hyperparameters for pretraining (left) and supervised fine-tuning (right).}
\label{tab:training_hyperparams}
\begin{minipage}[t]{0.48\textwidth}
\centering
\subcaption{Pretraining}
\label{tab:pt_hyperparams}
\small
\begin{tabular}{lcc}
\toprule
\textbf{Hyperparameter} & \textbf{240M} & \textbf{1B} \\
\midrule
Batch size & 256 & 256 \\
Sequence length & 4096 & 4096 \\
Training steps & 200,000 & 200,000 \\
Learning rate & 0.001 & 0.001 \\
LR schedule & Cosine & Cosine \\
Warmup steps & 2,000 & 2,000 \\
Weight decay & 0.001 & 0.001 \\
End decay & 0.1 & 0.1 \\
Gradient clipping & 1.0 & 1.0 \\
Adam $\beta_1$ & 0.9 & 0.9 \\
Adam $\beta_2$ & 0.95 & 0.95 \\
Precision & bfloat16 & bfloat16 \\
\bottomrule
\end{tabular}
\end{minipage}
\hfill
\begin{minipage}[t]{0.48\textwidth}
\centering
\subcaption{Supervised Fine-tuning}
\label{tab:sft_hyperparams}
\small
\begin{tabular}{lcc}
\toprule
\textbf{Hyperparameter} & \textbf{240M} & \textbf{1B} \\
\midrule
SFT dataset & \multicolumn{2}{c}{\texttt{tulu-v2-mix}} \\
Training Steps & 6,365 & 12,730 \\
Epochs & 5 & 10 \\
Batch size & 256 & 256 \\
Sequence length & 4096 & 4096 \\
Learning rate & 2e-5 & 1e-5 \\
LR schedule & Cosine & Cosine \\
Weight Decay & 0.001   & 0.001 \\
Warmup steps & 43 & 43 \\
Precision & bfloat16 & bfloat16 \\
\bottomrule
\end{tabular}
\end{minipage}
\end{table*}

\clearpage
\newpage
\subsection{Pretraining Data Mixtures}
\label{app:data_mixtures}

We construct 9 pretraining data mixtures by systematically varying two orthogonal dimensions: (1) the \textit{source} of web-crawled data, and (2) the \textit{proportion} of code versus web content.
This factorial design enables us to disentangle the effects of data quality from data composition.

\paragraph{Data Sources.}
Each mixture combines three categories of training data:
\begin{itemize}[leftmargin=1.5em]
    \item \textbf{Web-crawled data}: General web text from one of three sources---RefinedWeb~\citep{penedo2023refinedweb}, FineWeb-Edu~\citep{lozhkov2024fineweb-edu}, or DCLM~\citep{li2025datacomplmsearchgenerationtraining}. These sources differ in filtering strategies: RefinedWeb applies minimal quality filtering, FineWeb-Edu aggressively filters for educational content, and DCLM uses model-based quality scoring.
    \item \textbf{Code data}: Programming-related content from either StarCoder~\citep{li2023starcoder} or The Stack v2~\citep{lozhkov2024starcoder}, covering diverse programming languages and software documentation.
    \item \textbf{Curated knowledge}: High-quality reference material from RedPajama-v2~\citep{weber2024redpajama}, including Wikipedia, ArXiv papers, StackExchange discussions, and books.
\end{itemize}

\paragraph{Mixture Naming Convention.}
Each mixture is denoted \texttt{V$x$P$y$} where:
\begin{itemize}[leftmargin=1.5em]
    \item $x \in \{1, 2, 3\}$ indicates the web data source: V1 = RefinedWeb, V2 = FineWeb-Edu, V3 = DCLM
    \item $y \in \{0, 1, 2\}$ indicates the proportion configuration: P0 = web-heavy, P1 = balanced, P2 = code-heavy
\end{itemize}

\paragraph{Proportion Configurations.}
The P$y$ suffix controls the balance between data categories:
\begin{itemize}[leftmargin=1.5em]
    \item \textbf{P0 (Web-heavy)}: 65\% web, 25\% code, 10\% curated
    \item \textbf{P1 (Balanced)}: 45\% web, 35\% code, 20\% curated
    \item \textbf{P2 (Code-heavy)}: 25\% web, 45\% code, 30\% curated
\end{itemize}

Table~\ref{tab:data_mixtures} provides the complete specification of all 9 mixtures.

\begin{table}[ht!]
\centering
\caption{Pretraining data mixture compositions. All mixtures include curated knowledge sources from RedPajama-v2 (Wikipedia, ArXiv, StackExchange, Books) at the specified proportion.}
\label{tab:data_mixtures}
\begin{tabular}{lccc}
\toprule
\textbf{Mixture} & \textbf{Web (\%)} & \textbf{Code (\%)} & \textbf{Curated (\%)} \\
\midrule
V1P0 & RefinedWeb (65) & StarCoder (25) & 10 \\
V1P1 & RefinedWeb (45) & StarCoder (35) & 20 \\
V1P2 & RefinedWeb (25) & StarCoder (45) & 30 \\
\midrule
V2P0 & FineWeb-Edu (65) & Stack-v2 (25) & 10 \\
V2P1 & FineWeb-Edu (45) & Stack-v2 (35) & 20 \\
V2P2 & FineWeb-Edu (25) & Stack-v2 (45) & 30 \\
\midrule
V3P0 & DCLM (65) & Stack-v2 (25) & 10 \\
V3P1 & DCLM (45) & Stack-v2 (35) & 20 \\
V3P2 & DCLM (25) & Stack-v2 (45) & 30 \\
\bottomrule
\end{tabular}
\end{table}

This design enables systematic analysis of: (1) \textit{web source effects} by comparing V1 vs.\ V2 vs.\ V3 at fixed proportions, and (2) \textit{code proportion effects} by comparing P0 vs.\ P1 vs.\ P2 within each web source.

\clearpage
\newpage
\subsection{Evaluation Benchmarks}
\label{app:benchmarks}

We evaluate model performance across 20 benchmarks organized into 4 capability categories: \textit{Commonsense}, \textit{Science}, \textit{NLI}, and \textit{Semantic}.
Table~\ref{tab:benchmarks} provides complete benchmark details.

\begin{table*}[ht!]
\centering
\caption{Evaluation benchmarks organized by capability category. $|$Test$|$ indicates evaluation set size. All benchmarks are evaluated in a multiple-choice format using log-probability scoring.}
\label{tab:benchmarks}
\small
\begin{tabular}{llcp{4.5cm}}
\toprule
\textbf{Category} & \textbf{Benchmark} & \textbf{$|$Test$|$} & \textbf{Description} \\
\midrule
\multirow{7}{*}{\textcolor{Commonsense}{Commonsense}} 
& HellaSwag~\citep{zellers2019hellaswag} & 10,042 & Sentence completion requiring physical commonsense \\
& PIQA~\citep{bisk2020piqa} & 1,838 & Physical intuition about everyday objects and actions \\
& COPA~\citep{wang2019superglue} & 500 & Causal reasoning about everyday events \\
& SIQA~\citep{sap2019socialiqacommonsensereasoningsocial} & 1,954 & Social and emotional intelligence reasoning \\
& CommonsenseQA~\citep{talmor2019commonsenseqaquestionansweringchallenge} & 1,221 & General commonsense knowledge questions \\
& WinoGrande~\citep{sakaguchi2020winogrande} & 1,267 & Pronoun resolution requiring commonsense \\
& BoolQ~\citep{wang2019superglue} & 3,270 & Yes/no questions requiring world knowledge \\
\midrule
\multirow{4}{*}{\textcolor{Science}{Science}} 
& ARC-Challenge~\citep{clark2018arc} & 1,172 & Grade-school science questions (hard subset) \\
& ARC-Easy~\citep{clark2018arc} & 2,376 & Grade-school science questions (easy subset) \\
& SciQ~\citep{welbl2017crowdsourcingmultiplechoicescience} & 1,000 & Science exam questions with supporting passages \\
& OpenBookQA~\citep{mihaylov2018openbookqa} & 500 & Elementary science requiring external knowledge \\
\midrule
\multirow{4}{*}{\textcolor{NLI}{NLI}} 
& MNLI~\citep{williams2018mnli} & 9,815 & Multi-genre natural language inference \\
& QNLI~\citep{wang2019gluemultitaskbenchmarkanalysis} & 5,463 & Question-answering NLI (from SQuAD) \\
& RTE~\citep{wang2019superglue} & 277 & Recognizing textual entailment \\
& CB~\citep{wang2019superglue} & 56 & CommitmentBank textual entailment \\
\midrule
\multirow{5}{*}{\textcolor{Semantic}{Semantic}} 
& QQP~\citep{sharma2019qqp} & 40,430 & Quora question paraphrase detection \\
& MRPC~\citep{dolan2005mrpc} & 408 & Microsoft paraphrase corpus \\
& WiC~\citep{pilehvar2019wic} & 638 & Word-in-context sense disambiguation \\
& WSC~\citep{levesque2012wsc} & 104 & Winograd schema coreference resolution \\
& MultiRC~\citep{khashabi2018multirc} & 4,848 & Multi-sentence reading comprehension \\
\bottomrule
\end{tabular}
\end{table*}

\paragraph{Evaluation Protocol.}
All benchmarks are evaluated using a multiple-choice format with log-probability scoring.
For each example, we compute the log-probability of each candidate answer conditioned on the prompt and select the answer with the highest probability.
We report two metrics:
\begin{itemize}[leftmargin=1.5em]
    \item \textbf{Accuracy}: The proportion of examples where the model selects the correct answer.
    \item \textbf{Confidence}: The probability assigned to the selected answer (correct or incorrect), averaged across the evaluation set. This measures the model's certainty in its predictions.
\end{itemize}

\clearpage
\newpage
\section{Additional Results}
\label{app:results}
\subsection{Cross-Stage Correlation Results}
\label{app:cross_stage_results}

This section presents complete cross-benchmark correlation heatmaps for both accuracy and confidence metrics across training stages and model scales.

\subsubsection{Accuracy Correlations}

\begin{figure*}[ht!]
\centering
\includegraphics[width=0.75\textwidth]{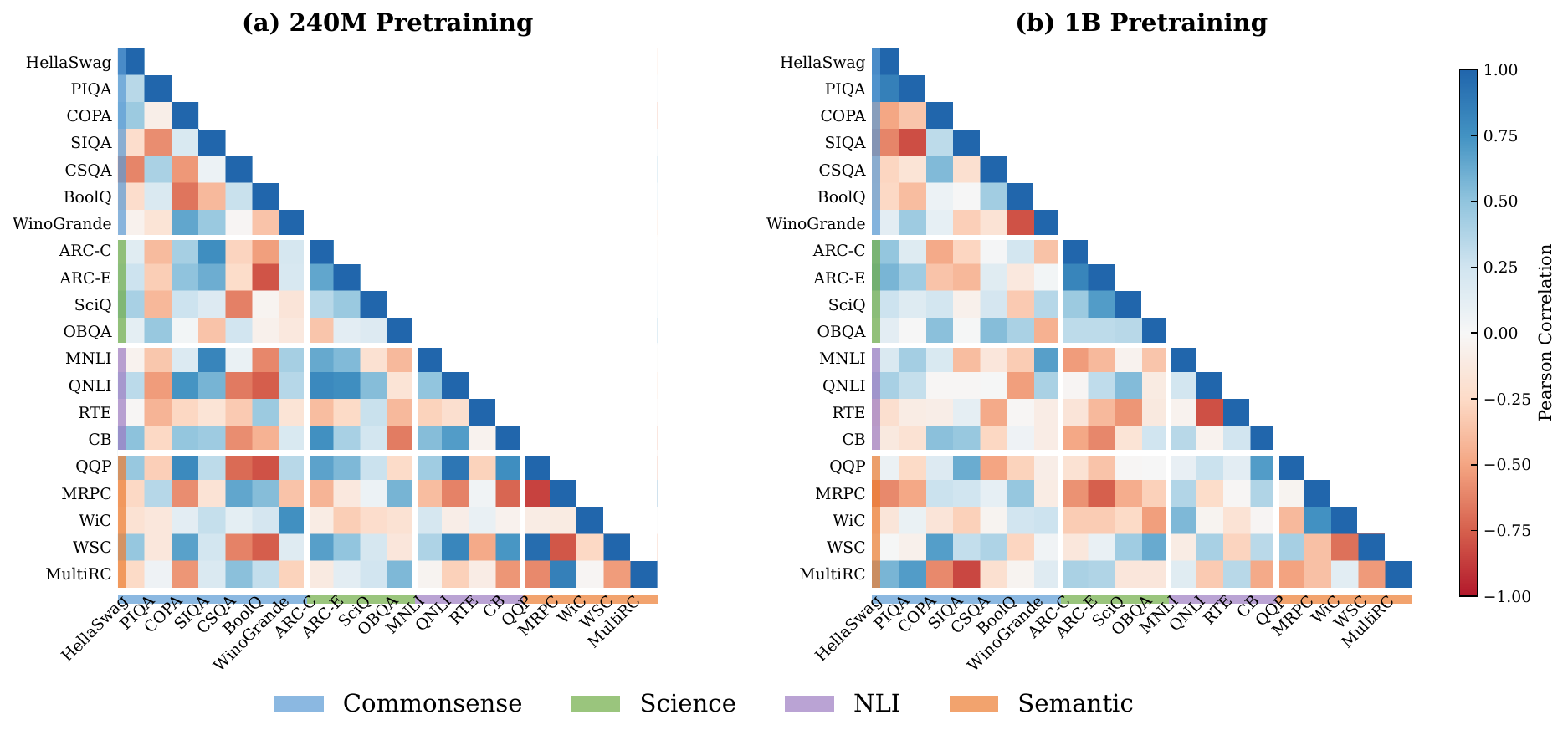}
\caption{\textbf{Cross-benchmark correlation during pretraining.} Each cell shows the Pearson correlation between two benchmarks' accuracy across data mixtures during pretraining only. The 1B model shows more positive within-category structure, while the 240M model exhibits competition (negative correlations) across categories.}
\label{fig:cross_benchmark_pt}
\end{figure*}

\begin{figure*}[ht!]
\centering
\includegraphics[width=0.7\textwidth]{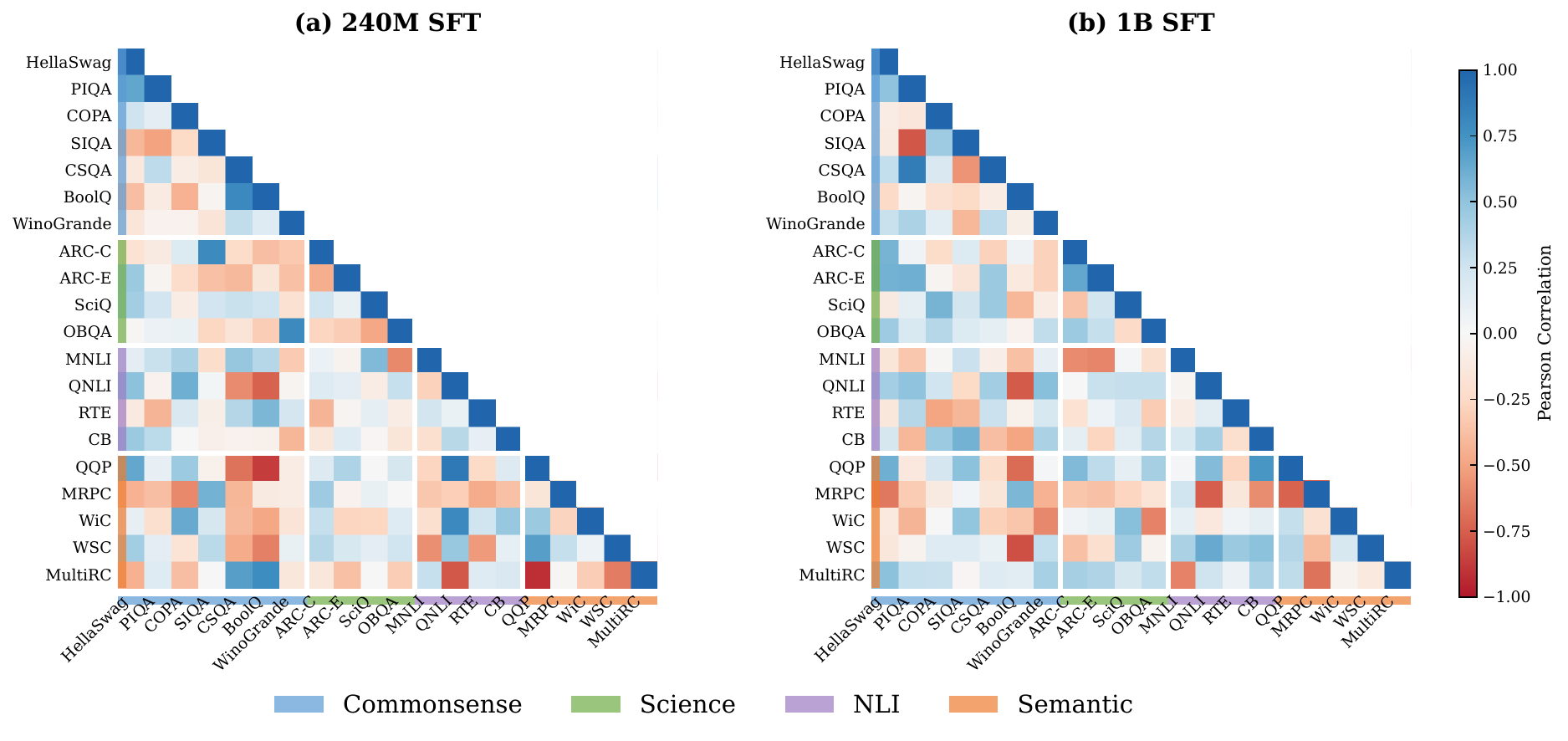}
\caption{\textbf{Cross-benchmark correlation during SFT.} Each cell shows the Pearson correlation between two benchmarks' accuracy across data mixtures after SFT. Patterns largely mirror pretraining, indicating that SFT preserves rather than reorganizes benchmark relationships.}
\label{fig:cross_benchmark_sft}
\end{figure*}

\begin{figure*}[ht!]
\centering
\includegraphics[width=0.7\textwidth]{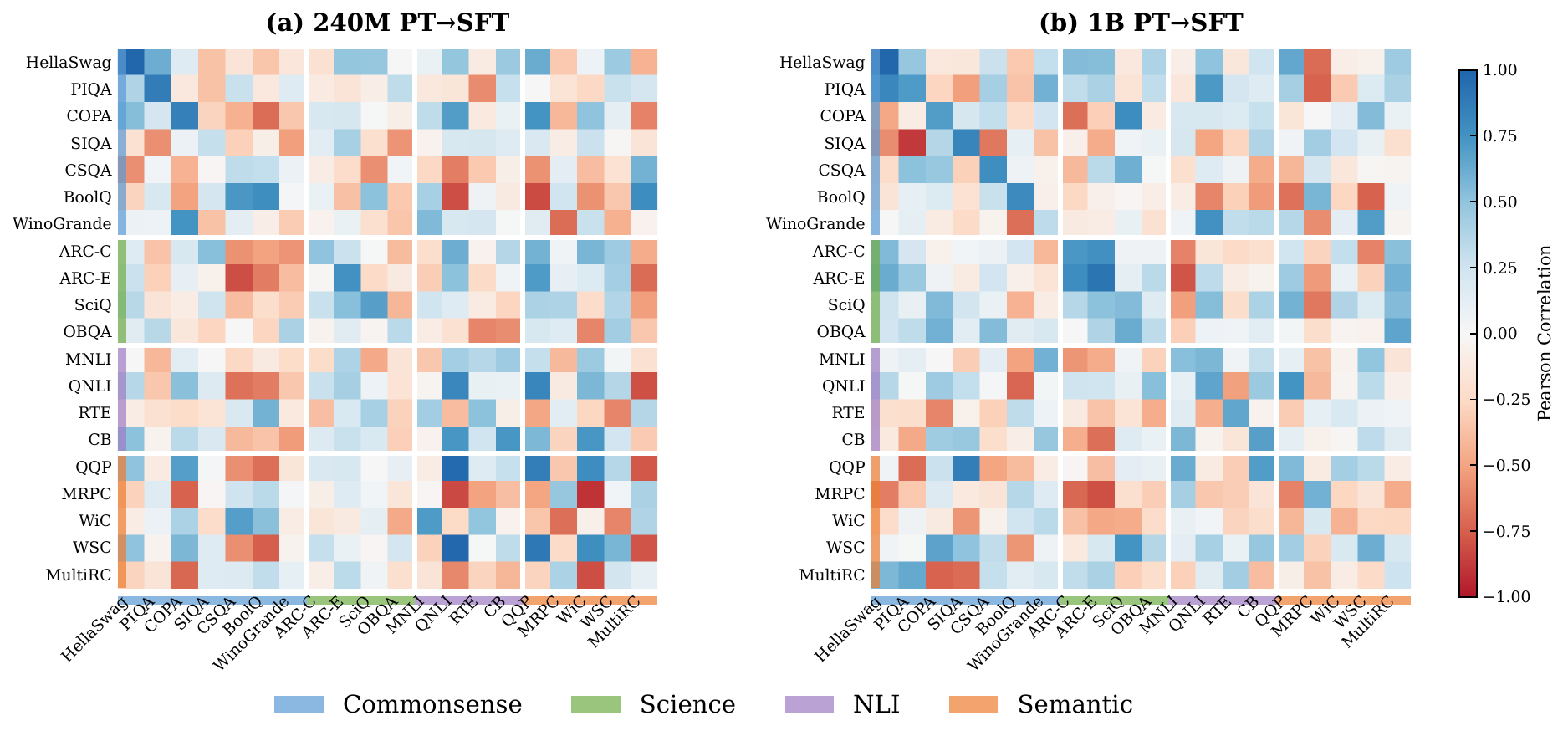}
\caption{\textbf{Cross-benchmark correlation for PT$\rightarrow$SFT transfer.} Each cell shows the Pearson correlation between one benchmark's PT accuracy and another's SFT accuracy. The diagonal represents same-benchmark transfer. Off-diagonal elements reveal cross-benchmark predictive relationships.}
\label{fig:cross_benchmark_pt_sft}
\end{figure*}

Figures~\ref{fig:cross_benchmark_pt}--\ref{fig:cross_benchmark_pt_sft} show pairwise benchmark correlations during pretraining, after SFT, and across stages (PT$\rightarrow$SFT).

\textbf{Key observations:} During pretraining (Figure~\ref{fig:cross_benchmark_pt}), the 240M model exhibits widespread negative correlations---particularly between Commonsense benchmarks and other categories---indicating that data mixtures optimizing one capability often degrade others. The 1B model shows more coherent positive structure, especially within the Science block, suggesting that larger models learn more generalizable representations that benefit multiple benchmarks simultaneously.

After SFT (Figure~\ref{fig:cross_benchmark_sft}), correlation patterns largely persist but with notable changes: the NLI block becomes more internally coherent at 1B, while cross-category relationships weaken at 240M. This suggests SFT primarily preserves rather than reorganizes the benchmark relationships established during pretraining.

The PT$\rightarrow$SFT transfer heatmap (Figure~\ref{fig:cross_benchmark_pt_sft}) reveals which pretraining benchmarks predict which SFT outcomes. HellaSwag emerges as a robust predictor across both scales, while WiC and MNLI show inconsistent or negative transfer, particularly at 240M.

\begin{figure*}[htbp!]
\centering
\includegraphics[width=0.7\textwidth]{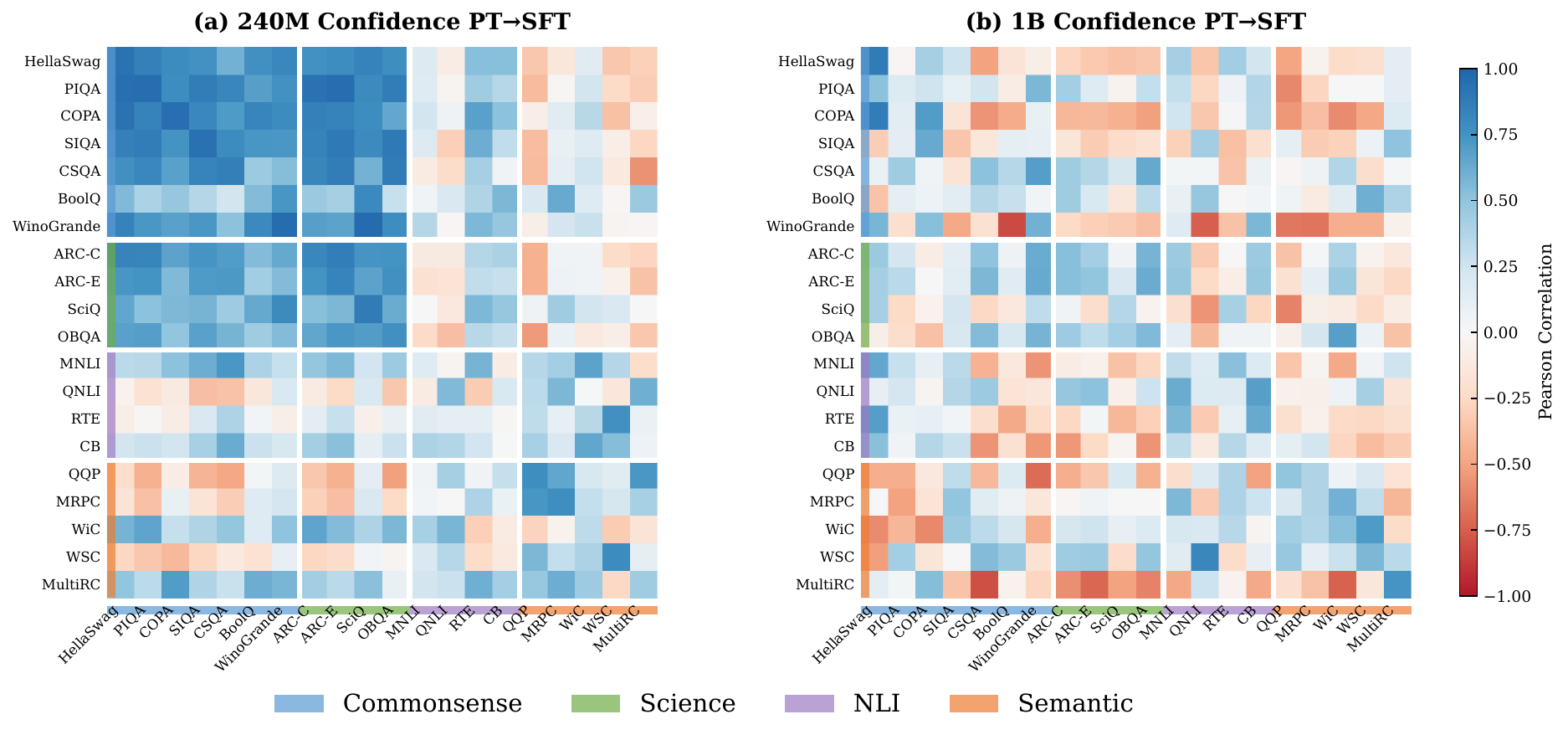}
\caption{\textbf{Cross-benchmark correlation for confidence PT$\rightarrow$SFT transfer.} Compared to accuracy (Figure~\ref{fig:cross_benchmark_pt_sft}), the 240M model shows substantially more positive correlations (blue), while 1B shows more heterogeneous patterns, indicating task-specific calibration development at scale.}
\label{fig:conf_heatmap}
\end{figure*}

\subsubsection{Confidence Correlations}

Figure~\ref{fig:conf_heatmap} presents the confidence correlation heatmap for PT$\rightarrow$SFT transfer, which can be compared with the accuracy version (Figure~\ref{fig:cross_benchmark_pt_sft}).

\textbf{Key observations:} The 240M model shows substantially more positive correlations (blue) for confidence than for accuracy, indicating that calibration patterns transfer more reliably than performance rankings at smaller scale. The confidence heatmap displays a more uniform positive structure, suggesting that a model's tendency to be confident or uncertain is a global property that persists across benchmarks and training stages.

In contrast, the 1B model shows more heterogeneous confidence patterns, with task-specific calibration that does not transfer as uniformly. This supports our main finding that larger models develop more nuanced, task-dependent uncertainty estimates rather than a single global calibration profile.

\clearpage
\newpage
\subsection{Per-Benchmark Transfer Statistics}
\label{app:per_benchmark}

Table~\ref{tab:per_benchmark_transfer} provides complete transfer statistics for all benchmarks, enabling practitioners to identify reliable early-stage evaluation candidates.

\textbf{Reliability tiers:} Based on our analysis, benchmarks can be grouped into reliability tiers. \textit{Highly reliable} (both accuracy and confidence $r > 0.6$ at both scales): HellaSwag, PIQA, COPA, QQP. \textit{Moderately reliable} (mixed patterns): ARC-E, SciQ, BoolQ, MRPC. \textit{Unreliable} ($r < 0.3$ or negative at either scale): WiC, WSC, MultiRC, WinoGrande (240M), MNLI (240M).

\textbf{Scale-dependent reliability:} Several benchmarks show dramatically different transfer at different scales. WinoGrande improves from $r=-0.31$ (240M) to $r=0.32$ (1B) for accuracy, while MNLI improves from $r=-0.34$ to $r=0.53$. This suggests that practitioners working with smaller proxy models should be particularly cautious about extrapolating from these benchmarks.

\begin{table}[ht!]
\centering
\caption{Per-benchmark cross-stage correlation (PT$\to$SFT) for accuracy and confidence. 
Benchmarks are grouped by category. 
Bold values indicate strong transfer ($r > 0.7$); \textcolor{red}{red} indicates negative transfer.}
\label{tab:per_benchmark_transfer}
\small
\begin{tabular}{ll|cc|cc}
\toprule
& & \multicolumn{2}{c|}{\textbf{Accuracy} $r_{\text{acc}}^{\text{stage}}$} & \multicolumn{2}{c}{\textbf{Confidence} $r_{\text{conf}}^{\text{stage}}$} \\
\textbf{Category} & \textbf{Benchmark} & \textbf{240M} & \textbf{1B} & \textbf{240M} & \textbf{1B} \\
\midrule
\multirow{7}{*}{\textcolor{Commonsense}{Commonsense}} 
& HellaSwag & \textbf{0.99} & \textbf{0.99} & \textbf{0.93} & \textbf{0.88} \\
& PIQA & \textbf{0.87} & \textbf{0.71} & \textbf{0.95} & 0.18 \\
& COPA & \textbf{0.85} & 0.70 & \textbf{0.94} & \textbf{0.70} \\
& SIQA & 0.30 & \textbf{0.82} & \textbf{0.93} & \textcolor{red}{-0.35} \\
& CommonsenseQA & 0.32 & \textbf{0.77} & \textbf{0.86} & 0.51 \\
& WinoGrande & \textcolor{red}{-0.31} & 0.32 & \textbf{0.95} & 0.59 \\
& BoolQ & \textbf{0.77} & \textbf{0.80} & 0.54 & 0.29 \\
\midrule
\multirow{4}{*}{\textcolor{Science}{Science}} 
& ARC-Challenge & 0.50 & \textbf{0.72} & \textbf{0.81} & 0.53 \\
& ARC-Easy & \textbf{0.76} & \textbf{0.91} & \textbf{0.83} & 0.50 \\
& SciQ & 0.69 & 0.54 & \textbf{0.88} & 0.36 \\
& OpenBookQA & 0.34 & 0.33 & \textbf{0.76} & 0.55 \\
\midrule
\multirow{4}{*}{\textcolor{NLI}{NLI}} 
& MNLI & \textcolor{red}{-0.34} & 0.53 & 0.16 & 0.30 \\
& QNLI & \textbf{0.81} & 0.66 & 0.55 & 0.18 \\
& RTE & 0.51 & 0.66 & 0.11 & 0.11 \\
& CB & \textbf{0.73} & 0.68 & 0.01 & 0.16 \\
\midrule
\multirow{5}{*}{\textcolor{Semantic}{Semantic}} 
& QQP & \textbf{0.87} & 0.55 & \textbf{0.77} & 0.49 \\
& MRPC & 0.48 & 0.60 & \textbf{0.77} & 0.38 \\
& WiC & \textcolor{red}{-0.07} & \textcolor{red}{-0.43} & 0.32 & 0.52 \\
& WSC & 0.57 & 0.61 & \textbf{0.78} & 0.57 \\
& MultiRC & 0.11 & 0.26 & 0.44 & \textbf{0.74} \\
\midrule
\multicolumn{2}{l|}{\textbf{Mean (all 20 benchmarks)}} & 0.49 & 0.59 & 0.66 & 0.41 \\
\bottomrule
\end{tabular}
\end{table}

\clearpage
\newpage
\subsection{Performance-Confidence Alignment}
\label{app:alignment}

We provide detailed statistics for the performance-confidence alignment analysis introduced in Section~\ref{sec:alignment_protocol}.
Performance-confidence alignment measures the correlation between a model's accuracy and its confidence across benchmarks, quantifying whether models are appropriately certain when correct and uncertain when incorrect.

\subsubsection{Definition and Measurement Protocol}

For a given model configuration (scale, stage, data mixture), we compute the \textbf{performance-confidence alignment} as:
\begin{equation}
    r_{\text{align}} = \text{corr}(\mathbf{a}, \mathbf{c})
\end{equation}
where $\mathbf{a}$ and $\mathbf{c}$ denote accuracy and confidence vectors across benchmarks within a capability category.
High positive $r_{\text{align}}$ indicates well-calibrated models (confident when correct); negative values indicate miscalibration (overconfident on errors).

We report $r_{\text{align}}$ at two granularities:
\begin{itemize}[leftmargin=1.5em, itemsep=0.05em]
    \item \textbf{Capacity Category}: Correlation computed across all benchmarks within each capability category, averaged over data mixture configurations.
    \item \textbf{Data Configuration}: Correlation computed for each unique combination of web source and code proportion, enabling analysis of data mixture effects.
\end{itemize}

\subsubsection{Category-Level Alignment Statistics}

Table~\ref{tab:alignment_category_summary} presents the category-level performance-confidence alignment across scales and training stages.

\begin{table}[ht!]
\centering
\caption{\textbf{Performance-confidence alignment by capability category.} Values show $r_{\text{align}}$ computed over all data mixtures. Science exhibits consistently high alignment ($r > 0.7$), while Commonsense and Semantic show weak or negative alignment indicating systematic miscalibration.}
\label{tab:alignment_category_summary}
\small
\begin{tabular}{llcccc}
\toprule
\textbf{Scale} & \textbf{Stage} & \textbf{Commonsense} & \textbf{Science} & \textbf{NLI} & \textbf{Semantic} \\
\midrule
\multirow{2}{*}{240M} 
& PT  & $0.05$ & $0.75$ & $0.35$ & $-0.18$ \\
& SFT & $0.04$ & $0.80$ & $0.25$ & $-0.37$ \\
\midrule
\multirow{2}{*}{1B} 
& PT  & $-0.12$ & $0.81$ & $-0.12$ & $-0.06$ \\
& SFT & $-0.16$ & $0.83$ & $0.39$ & $-0.16$ \\
\bottomrule
\end{tabular}
\end{table}

\paragraph{Key Observations.}
\begin{enumerate}[leftmargin=1.5em, itemsep=0.05em]
    \item \textbf{Science exhibits strong positive alignment} ($r_{\text{align}} \approx 0.75$--$0.83$) across all configurations, indicating that models are appropriately calibrated on scientific reasoning tasks.
    
    \item \textbf{Commonsense shows weak alignment} near zero at 240M and slightly negative at 1B ($r_{\text{align}} \approx -0.12$ to $-0.16$), suggesting models are neither well-calibrated nor systematically miscalibrated.
    
    \item \textbf{Semantic exhibits consistent negative alignment} ($r_{\text{align}} \approx -0.18$ to $-0.37$ at 240M), indicating overconfidence on incorrect predictions for paraphrase and semantic similarity tasks.
    
    \item \textbf{NLI alignment varies with scale and stage}: positive at 240M PT ($r_{\text{align}} = 0.35$) but negative at 1B PT ($r_{\text{align}} = -0.12$), with SFT generally improving NLI calibration at 1B.
\end{enumerate}


\subsubsection{Scale-Dependent Effects of Educational Filtering}
\label{app:alignment_scale_dependent}

\begin{figure}[ht!]
\centering
\includegraphics[width=0.7\textwidth]{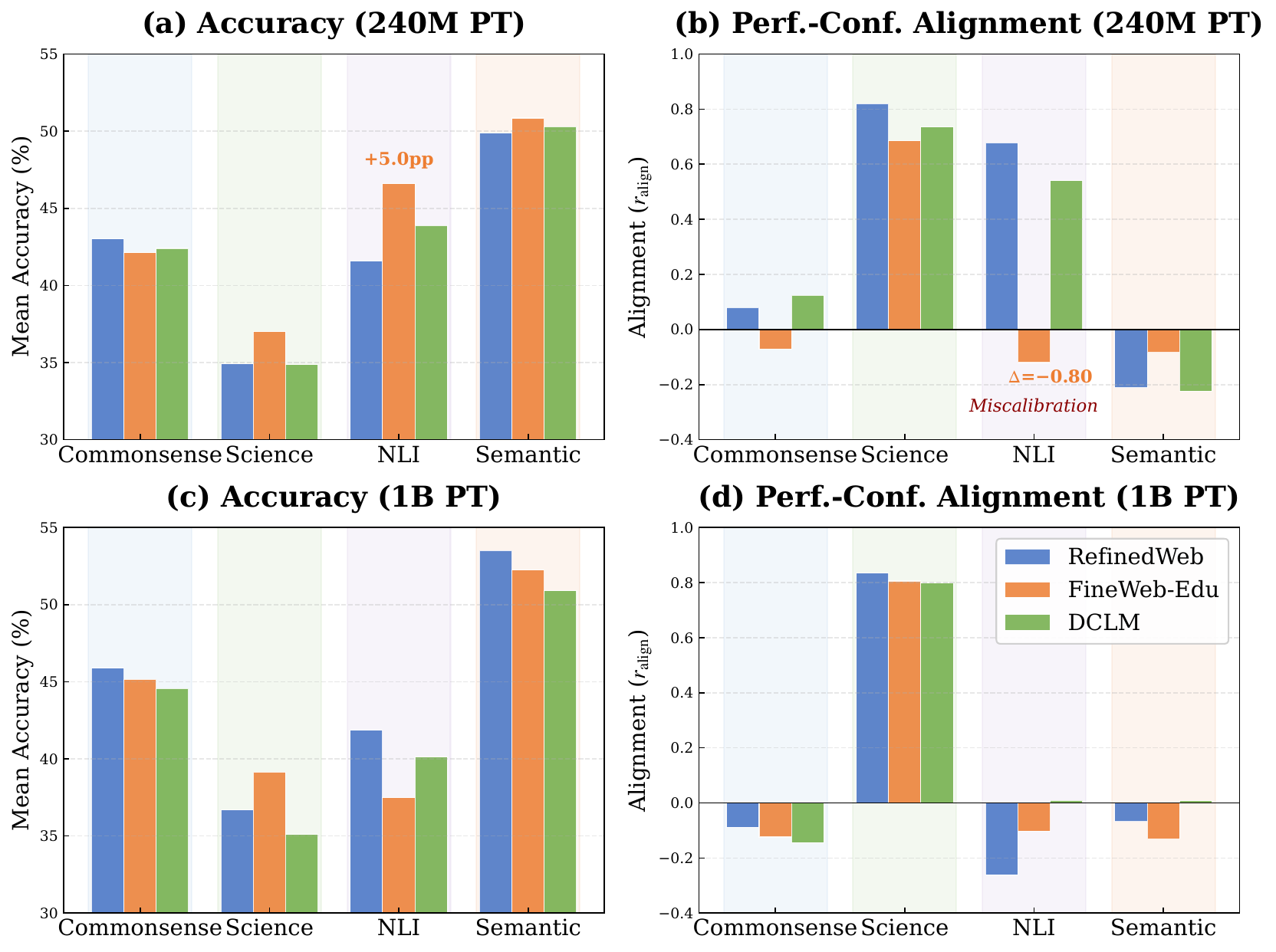}
\caption{\textbf{Educational filtering effects are scale-dependent.}
\textbf{Top row (240M PT)}: (a)~FineWeb-Edu \emph{improves} NLI accuracy by +5.0pp compared to RefinedWeb; (b)~yet it \emph{severely degrades} NLI alignment ($\Delta r_{\text{align}} = -0.80$), flipping from well-calibrated ($r = 0.68$) to miscalibrated ($r = -0.12$).
\textbf{Bottom row (1B PT)}: (c)~The accuracy effect \emph{reverses}---FineWeb-Edu now \emph{degrades} NLI accuracy relative to RefinedWeb; (d)~alignment differences across web sources diminish substantially.
Science alignment remains robust across all configurations ($r_{\text{align}} \in [0.68, 0.83]$).
These scale-dependent dynamics challenge the practice of using small proxy models for data curation decisions.}
\label{fig:web_source_accuracy_alignment_appendix}
\end{figure}

A key finding from our analysis is that the effects of educational content filtering (FineWeb-Edu) on both accuracy and alignment are \emph{strongly scale-dependent}.
Figure~\ref{fig:web_source_accuracy_alignment_appendix} presents the category-level accuracy and alignment across different web sources at both 240M and 1B scales, revealing striking scale-dependent patterns.

\paragraph{Scale-Dependent Accuracy-Calibration Trade-offs.}
At \textbf{240M scale} (Figure~\ref{fig:web_source_accuracy_alignment_appendix}a--b), FineWeb-Edu exhibits a striking accuracy-calibration dissociation on NLI tasks:
\begin{itemize}[leftmargin=1.5em, itemsep=0.05em]
    \item \textbf{Accuracy improvement}: FineWeb-Edu yields +5.0pp higher NLI accuracy than RefinedWeb, suggesting that educational content filtering captures linguistic patterns beneficial for inference tasks.
    \item \textbf{Calibration degradation}: Despite accuracy gains, NLI alignment drops from $r_{\text{align}} = 0.68$ (RefinedWeb) to $r_{\text{align}} = -0.12$ (FineWeb-Edu)---a $\Delta = -0.80$ decrease that flips the model from well-calibrated to systematically overconfident on incorrect predictions.
\end{itemize}

At \textbf{1B scale} (Figure~\ref{fig:web_source_accuracy_alignment_appendix}c--d), these patterns \emph{reverse}:
\begin{itemize}[leftmargin=1.5em, itemsep=0.05em]
    \item \textbf{Accuracy reversal}: FineWeb-Edu now \emph{degrades} NLI accuracy by $-4.4$pp compared to RefinedWeb, indicating that the beneficial patterns captured at 240M do not transfer to larger models.
    \item \textbf{Alignment convergence}: The dramatic alignment differences observed at 240M largely disappear---all web sources produce similar (weakly negative) NLI alignment at 1B PT.
\end{itemize}

\paragraph{Robust Categories.}
In contrast to NLI's scale-dependent behavior, \textcolor{Science}{\textit{Science}} alignment remains consistently high ($r_{\text{align}} \approx 0.7$--$0.8$) across all web sources and both scales, suggesting that educational filtering preserves the structured factual knowledge essential for well-calibrated scientific reasoning.

\begin{figure}[ht!]
\centering
\includegraphics[width=0.8\textwidth]{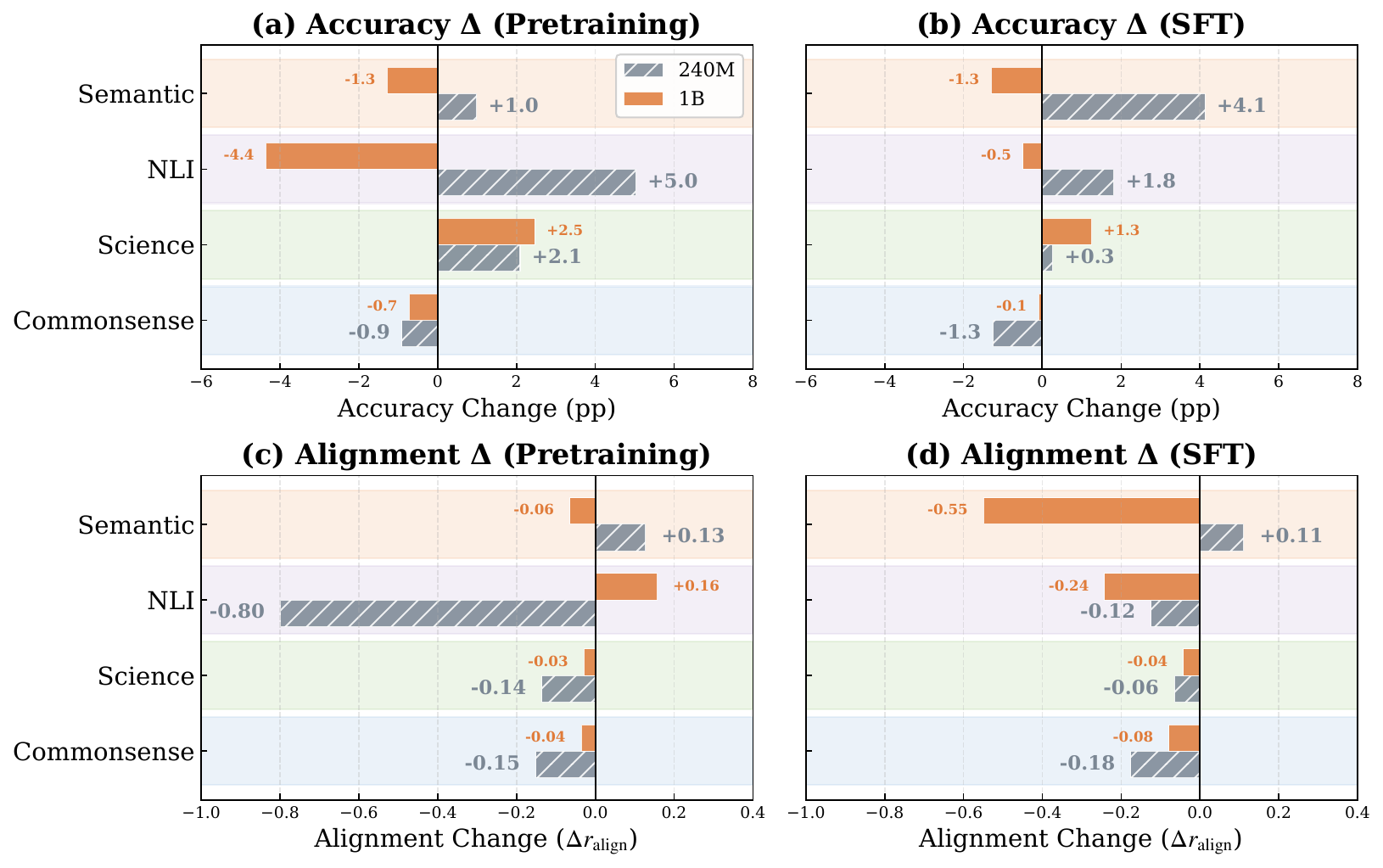}
\caption{\textbf{FineWeb-Edu vs.\ RefinedWeb: Category-level effects at PT and SFT stages.}
(a,b)~Accuracy changes; (c,d)~Alignment changes.
Left column: Pretraining; Right column: SFT.
The NLI accuracy effect \emph{reverses} between scales: FineWeb-Edu improves NLI by +5.0pp at 240M but \emph{degrades} it by $-4.4$pp at 1B.
Alignment effects also show scale-dependent patterns, with the severe 240M degradation ($\Delta = -0.80$) not replicating at 1B ($\Delta = +0.16$).}
\label{fig:fineweb_edu_delta_appendix}
\end{figure}

Figure~\ref{fig:fineweb_edu_delta_appendix} provides a complementary view, showing the FineWeb-Edu vs.\ RefinedWeb deltas across both scales and training stages.

Table~\ref{tab:fineweb_edu_delta_summary} summarizes the FineWeb-Edu vs.\ RefinedWeb comparison across capability categories.

\begin{table}[ht!]
\centering
\caption{\textbf{FineWeb-Edu vs.\ RefinedWeb: Category-level effects at PT stage.}
Accuracy ($\Delta$acc) in percentage points; Alignment ($\Delta r_{\text{align}}$) as correlation change.
NLI exhibits complete reversal between scales---improving at 240M but degrading at 1B.}
\label{tab:fineweb_edu_delta_summary}
\small
\begin{tabular}{l|cc|cc}
\toprule
\multirow{2}{*}{\textbf{Category}} & \multicolumn{2}{c|}{\textbf{Accuracy $\Delta$ (pp)}} & \multicolumn{2}{c}{\textbf{Alignment $\Delta$}} \\
& 240M & 1B & 240M & 1B \\
\midrule
\textcolor{Commonsense}{Commonsense} & $-0.9$ & $-0.7$ & $-0.15$ & $-0.04$ \\
\textcolor{Science}{Science} & $+2.1$ & $+2.5$ & $-0.14$ & $-0.03$ \\
\textcolor{NLI}{NLI} & $\mathbf{+5.0}$ & $\mathbf{-4.4}$ & $\mathbf{-0.80}$ & $\mathbf{+0.16}$ \\
\textcolor{Semantic}{Semantic} & $+1.0$ & $-1.3$ & $+0.13$ & $-0.06$ \\
\bottomrule
\end{tabular}
\end{table}

\paragraph{Scale-Dependent Patterns.}
The NLI category exhibits the most dramatic scale-dependent behavior:
\begin{itemize}[leftmargin=1.5em, itemsep=0.05em]
    \item \textbf{At 240M}: FineWeb-Edu \emph{improves} NLI accuracy by +5.0pp while \emph{severely degrading} alignment ($\Delta = -0.80$), flipping from well-calibrated ($r_{\text{align}} = 0.68$) to miscalibrated ($r_{\text{align}} = -0.12$).
    
    \item \textbf{At 1B}: The pattern \emph{reverses}---FineWeb-Edu \emph{degrades} NLI accuracy by $-4.4$pp while \emph{slightly improving} alignment ($\Delta = +0.16$).
    
    \item \textbf{Science is robust}: Science shows consistent accuracy improvement at both scales (+2.1pp at 240M, +2.5pp at 1B) with minimal alignment degradation, suggesting that educational filtering preserves structured factual knowledge beneficial for scientific reasoning.
\end{itemize}

\paragraph{Implications for Proxy Model Experiments.}
These scale-dependent effects challenge the common practice of using small proxy models for data curation decisions~\citep{kaplan2020scaling,hoffmann2022training,xie2023doremi}.
Data mixture choices that appear beneficial at 240M may produce opposite effects at 1B, particularly for linguistically nuanced tasks like NLI.
Practitioners should validate data curation decisions at multiple scales before committing to production training.

\subsubsection{Alignment by Web Data Source}

Table~\ref{tab:alignment_web_source_full} presents the complete performance-confidence alignment statistics broken down by web data source.

\begin{table}[ht!]
\centering
\caption{\textbf{Performance-confidence alignment by web data source} ($r_{\text{align}}$). FineWeb-Edu shows dramatically degraded NLI alignment at 240M PT ($-0.12$ vs.\ $0.68$ for RefinedWeb), but this effect diminishes at 1B.}
\label{tab:alignment_web_source_full}
\small
\begin{tabular}{lllcccc}
\toprule
\textbf{Scale} & \textbf{Stage} & \textbf{Web Source} & \textbf{Commonsense} & \textbf{Science} & \textbf{NLI} & \textbf{Semantic} \\
\midrule
\multirow{6}{*}{240M} 
& \multirow{3}{*}{PT}
& RefinedWeb  & $0.08$  & $0.82$ & $0.68$  & $-0.21$ \\
& & FineWeb-Edu & $-0.07$ & $0.68$ & $-0.12$ & $-0.08$ \\
& & DCLM        & $0.12$  & $0.73$ & $0.54$  & $-0.22$ \\
\cmidrule{2-7}
& \multirow{3}{*}{SFT}
& RefinedWeb  & $0.11$  & $0.83$ & $0.43$  & $-0.42$ \\
& & FineWeb-Edu & $-0.07$ & $0.77$ & $0.31$  & $-0.31$ \\
& & DCLM        & $0.08$  & $0.81$ & $0.09$  & $-0.34$ \\
\midrule
\multirow{6}{*}{1B} 
& \multirow{3}{*}{PT}
& RefinedWeb  & $-0.09$ & $0.83$ & $-0.26$ & $-0.07$ \\
& & FineWeb-Edu & $-0.12$ & $0.81$ & $-0.10$ & $-0.13$ \\
& & DCLM        & $-0.14$ & $0.80$ & $0.01$  & $0.01$ \\
\cmidrule{2-7}
& \multirow{3}{*}{SFT}
& RefinedWeb  & $-0.15$ & $0.85$ & $0.66$  & $0.11$ \\
& & FineWeb-Edu & $-0.23$ & $0.81$ & $0.42$  & $-0.44$ \\
& & DCLM        & $-0.10$ & $0.82$ & $0.32$  & $-0.15$ \\
\bottomrule
\end{tabular}
\end{table}

\paragraph{FineWeb-Edu Effect on NLI Calibration.}
The most striking finding is the severe degradation of NLI alignment under FineWeb-Edu at 240M:
\begin{itemize}[leftmargin=1.5em, itemsep=0.05em]
    \item \textbf{PT stage (240M)}: $r_{\text{align}}$ drops from $0.68$ (RefinedWeb) to $-0.12$ (FineWeb-Edu), a decrease of $\Delta = -0.80$.
    \item \textbf{SFT stage}: The gap narrows ($0.43$ vs.\ $0.31$), but FineWeb-Edu still yields lower NLI alignment.
    \item \textbf{At 1B}: The effect diminishes substantially---FineWeb-Edu actually shows \emph{better} NLI alignment than RefinedWeb at PT ($-0.10$ vs.\ $-0.26$).
    \item \textbf{At 1B SFT}: FineWeb-Edu produces notably negative Semantic alignment ($r_{\text{align}} = -0.44$) compared to RefinedWeb ($0.11$), suggesting scale-dependent effects shift across categories.
\end{itemize}

This suggests that educational content filtering removes diverse linguistic patterns that are essential for well-calibrated natural language inference at smaller scales, while the effect diminishes or reverses at larger scales where models may develop more robust representations.

\subsubsection{Alignment by Code Proportion}

Table~\ref{tab:alignment_code_prop_full} presents the performance-confidence alignment statistics broken down by code proportion.

\begin{table}[ht!]
\centering
\caption{\textbf{Performance-confidence alignment by code proportion} ($r_{\text{align}}$, averaged over web sources). At 1B, higher code proportion (45\%) improves Commonsense and NLI alignment despite degrading accuracy.}
\label{tab:alignment_code_prop_full}
\small
\begin{tabular}{lllcccc}
\toprule
\textbf{Scale} & \textbf{Stage} & \textbf{Code \%} & \textbf{Commonsense} & \textbf{Science} & \textbf{NLI} & \textbf{Semantic} \\
\midrule
\multirow{6}{*}{240M} 
& \multirow{3}{*}{PT}
& 25\% & $0.06$  & $0.79$ & $0.28$  & $-0.18$ \\
& & 35\% & $-0.01$ & $0.72$ & $0.66$  & $-0.31$ \\
& & 45\% & $0.05$  & $0.79$ & $0.31$  & $-0.07$ \\
\cmidrule{2-7}
& \multirow{3}{*}{SFT}
& 25\% & $0.08$  & $0.85$ & $0.35$  & $-0.28$ \\
& & 35\% & $-0.08$ & $0.80$ & $0.12$  & $-0.48$ \\
& & 45\% & $0.10$  & $0.82$ & $0.67$  & $-0.31$ \\
\midrule
\multirow{6}{*}{1B} 
& \multirow{3}{*}{PT}
& 25\% & $-0.21$ & $0.83$ & $-0.32$ & $0.07$ \\
& & 35\% & $-0.18$ & $0.81$ & $-0.64$ & $-0.25$ \\
& & 45\% & $-0.00$ & $0.81$ & $0.26$  & $0.14$ \\
\cmidrule{2-7}
& \multirow{3}{*}{SFT}
& 25\% & $-0.26$ & $0.85$ & $0.19$  & $-0.01$ \\
& & 35\% & $-0.12$ & $0.84$ & $0.71$  & $-0.17$ \\
& & 45\% & $-0.14$ & $0.82$ & $0.06$  & $-0.14$ \\
\bottomrule
\end{tabular}
\end{table}

\paragraph{Code Proportion Effects.}
Unlike web source effects, code proportion exhibits weaker but qualitatively distinct patterns:
\begin{itemize}[leftmargin=1.5em, itemsep=0.05em]
    \item \textbf{Accuracy-calibration dissociation at 1B}: Increasing code from 25\% to 45\% improves Commonsense alignment ($\Delta = +0.21$) and NLI alignment ($\Delta = +0.58$), despite degrading accuracy on these tasks.
    
    \item \textbf{Non-monotonic NLI pattern}: At 1B PT, 35\% code yields notably worse NLI alignment ($r_{\text{align}} = -0.64$) compared to both 25\% ($-0.32$) and 45\% ($+0.26$).
    
    \item \textbf{Science robustness}: Science alignment remains stable across all code proportions ($r_{\text{align}} \in [0.72, 0.85]$).
\end{itemize}

\subsubsection{Alignment Persistence Across Training Stages}

To assess whether performance-confidence alignment established during pretraining persists through SFT, we compute the correlation between PT and SFT alignment values across data mixtures for each category.

\begin{table}[ht!]
\centering
\caption{\textbf{Alignment persistence} (Pearson correlation between PT and SFT $r_{\text{align}}$ values across 9 data mixtures). Positive values indicate that alignment patterns established during pretraining persist through SFT; negative values indicate reorganization.}
\label{tab:alignment_persistence}
\small
\begin{tabular}{lcccc}
\toprule
\textbf{Scale} & \textbf{Commonsense} & \textbf{Science} & \textbf{NLI} & \textbf{Semantic} \\
\midrule
240M & $0.73$ & $0.62$ & $-0.41$ & $0.50$ \\
1B   & $0.54$ & $0.31$ & $-0.40$ & $0.27$ \\
\bottomrule
\end{tabular}
\end{table}

\paragraph{Persistence Patterns.}
Commonsense and Science categories show moderate-to-strong PT$\to$SFT alignment persistence ($r = 0.54$--$0.73$), indicating that calibration quality established during pretraining partially carries through supervised fine-tuning.
Notably, NLI shows \emph{negative} persistence ($r \approx -0.40$ at both scales), indicating that SFT \emph{reverses} rather than preserves the calibration patterns from pretraining---data mixtures that yield well-calibrated NLI during pretraining tend to produce poorly calibrated NLI after SFT, and vice versa.
This suggests that NLI calibration undergoes fundamental reorganization during instruction tuning, in contrast to other categories where pretraining calibration largely persists.

\subsubsection{Detailed Practical Implications}

The performance-confidence alignment analysis yields several actionable insights:

\begin{enumerate}[leftmargin=1.5em, itemsep=0.1em]
    \item \textbf{Science tasks are reliably well-calibrated}: Across all data mixtures and scales, Science benchmarks exhibit strong positive alignment ($r_{\text{align}} > 0.7$), making confidence scores trustworthy for scientific reasoning applications.
    
    \item \textbf{Commonsense and Semantic tasks require calibration intervention}: Weak or negative alignment on these categories suggests that post-hoc calibration techniques (e.g., temperature scaling) may be necessary for deployments requiring reliable uncertainty estimates.
    
    \item \textbf{Educational filtering effects are scale-dependent}: FineWeb-Edu produces dramatically different effects at 240M vs.\ 1B scales (Table~\ref{tab:fineweb_edu_delta_summary}). Practitioners should validate data curation decisions at multiple scales before production training.
    
    \item \textbf{NLI calibration reorganizes during SFT}: The negative alignment persistence for NLI ($r \approx -0.40$) indicates that pretraining calibration quality does not predict post-SFT calibration. Task-specific calibration techniques may be necessary for NLI applications.
    
    \item \textbf{Code proportion creates accuracy-calibration trade-offs}: Higher code proportions may improve calibration while degrading accuracy, offering a design choice depending on whether raw performance or reliable uncertainty is prioritized.
\end{enumerate}

\clearpage
\newpage
\subsection{Impact of Pretraining Data Mixture}
\label{app:data_mixture}

\subsubsection{Category-Level Code Effects}

\begin{figure}[ht!]
\centering
\includegraphics[width=0.75\columnwidth]{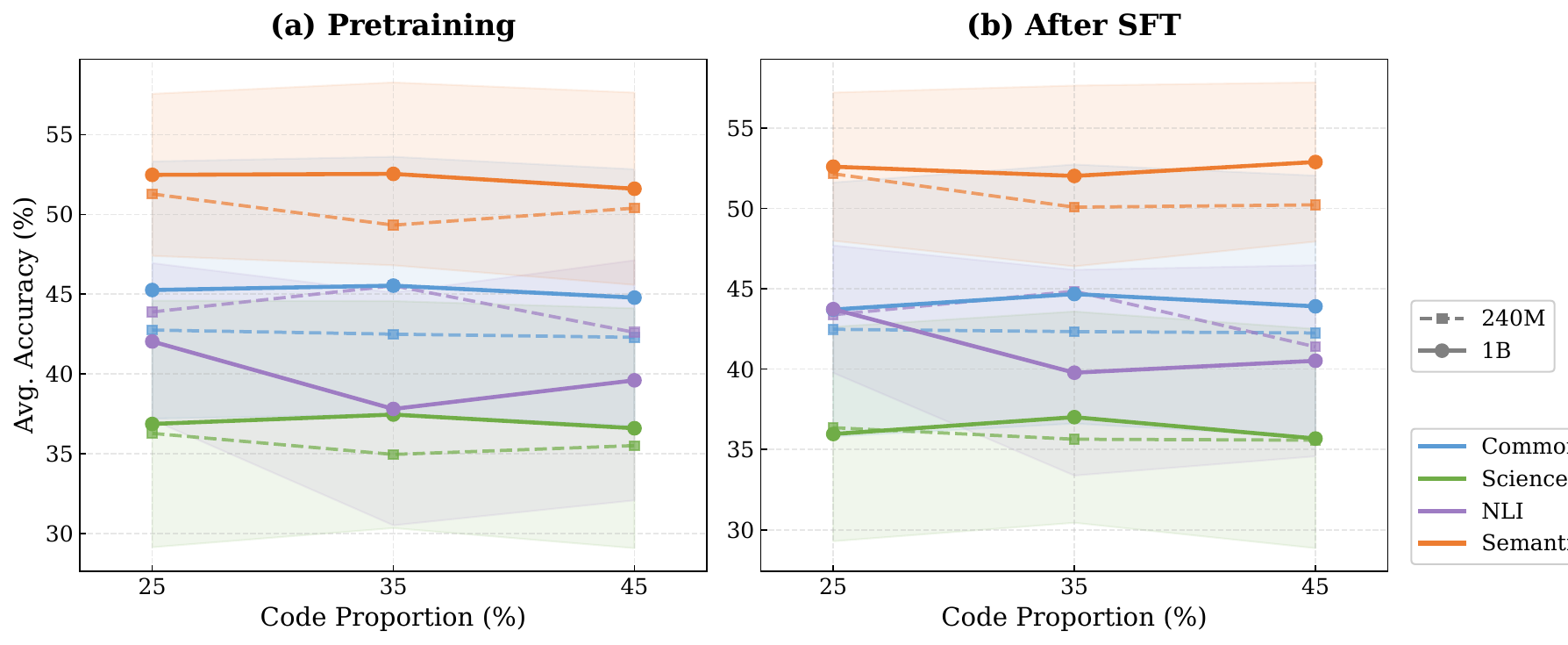}
\caption{\textbf{Category-level correlation between code proportion and accuracy.}
Each bar shows the Pearson correlation between code data proportion (25\%, 35\%, 45\%) and mean category accuracy across data mixtures.
\textbf{(a) 240M}: \textcolor{Commonsense}{Commonsense} shows strong negative correlation in both PT ($r = -0.86$) and SFT ($r = -0.78$).
\textbf{(b) 1B}: \textcolor{Commonsense}{Commonsense} remains negatively correlated ($r = -0.98$ PT), while \textcolor{Science}{Science} shifts to positive correlation ($r = +0.73$ SFT), suggesting scale-dependent interactions.}
\label{fig:category_code_effect}
\end{figure}

Figure~\ref{fig:category_code_effect} shows the correlation between code data proportion and category-level accuracy across both training stages and model scales.

\clearpage
\newpage
\subsubsection{Benchmark-Level Code Effects}

\begin{figure*}[ht!]
\centering
\includegraphics[width=\textwidth]{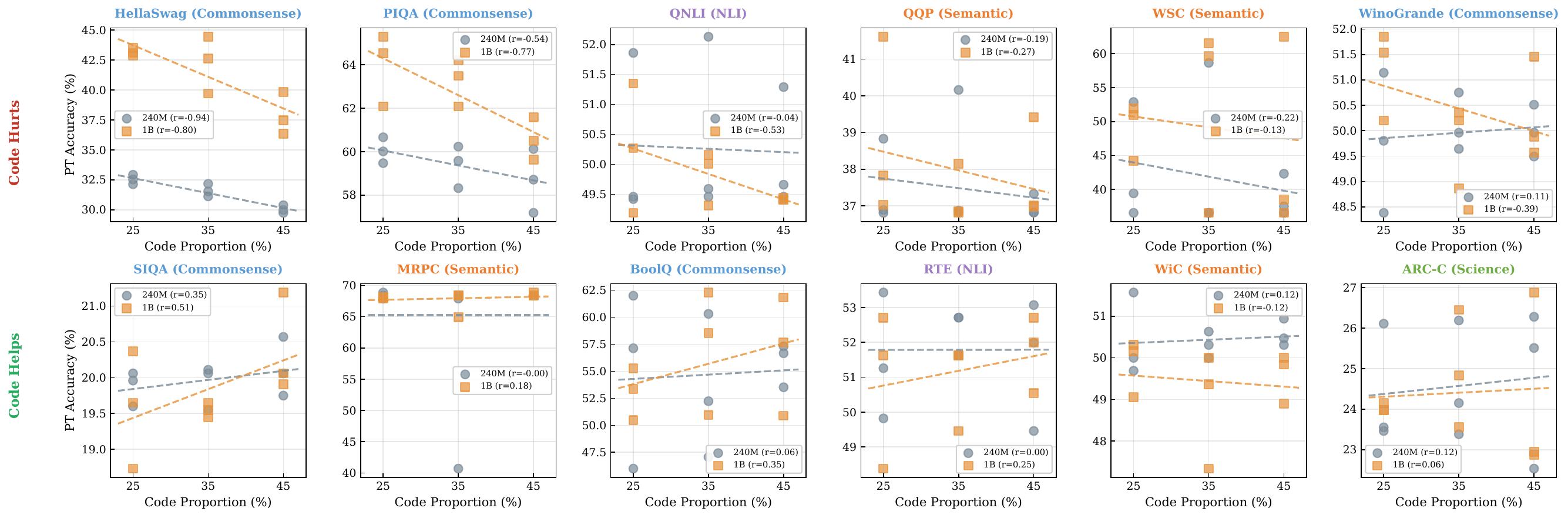}
\caption{\textbf{Benchmark-level sensitivity to code data proportion during pretraining.}
Each subplot shows the relationship between code proportion (25\%, 35\%, 45\%) and PT accuracy across 9 data mixtures.
\textbf{Top row (Code Hurts)}: HellaSwag shows the strongest degradation ($r = -0.94$ at 240M, $r = -0.80$ at 1B), followed by PIQA ($r = -0.77$ at 1B).
\textbf{Bottom row (Code Helps)}: Some other tasks shows positive effects with the increasing of code data proportion, suggesting potential reasoning benefits from code data on scientific reasoning (\textcolor{Science}{ARC-C}) and social reasoning (\textcolor{Commonsense}{SIQA})---in contrast to physical reasoning.
Title colors indicate category membership; dashed lines show linear trends.}
\label{fig:code_effect_extremes}
\end{figure*}

Figure~\ref{fig:code_effect_extremes} reveals that code data has highly heterogeneous effects across benchmarks---even within the same semantic category.

\textbf{Physical vs.\ social reasoning:} Within Commonsense, physical reasoning tasks (HellaSwag, PIQA) show strong negative correlations with code proportion ($r=-0.94$, $-0.77$ at 1B), while social reasoning tasks (SIQA, CommonsenseQA, BoolQ) show positive correlations ($r=0.51$, $0.24$, $0.35$ at 1B). This suggests that code data---which lacks descriptions of physical world interactions---may degrade grounded physical reasoning while benefiting more abstract logical reasoning.

\textbf{NLI heterogeneity:} The NLI category shows similarly mixed patterns: QNLI and MNLI are negatively affected by code at 1B ($r=-0.53$, $-0.68$), while RTE shows a modest positive effect ($r=0.25$). This may reflect differences in task formulation---QNLI and MNLI require nuanced language understanding that code data does not provide, while RTE's simpler binary classification may benefit from code's logical structure.

\textbf{Scale interactions:} Several benchmarks show opposite code effects at different scales. MNLI shifts from $r=0.12$ (240M) to $r=-0.68$ (1B), while RTE shifts from $r=0.00$ to $r=0.25$. This underscores that conclusions about data mixture effects from small-scale experiments may not extrapolate to larger models.

\clearpage
\newpage
\subsubsection{Web Source Comparison}

Table~\ref{tab:web_source_effects} summarizes how different web data sources affect category-level performance, averaged across code proportions.

\textbf{Key patterns:} FineWeb-Edu shows a slight advantage for Science benchmarks (+0.8pp over RefinedWeb), consistent with its educational content filtering. However, this advantage does not extend to Commonsense or NLI, where general web data performs comparably or better. DCLM shows competitive performance across categories, suggesting that model-based quality filtering can match educational filtering without domain restrictions.

\begin{table}[ht!]
\centering
\caption{Mean accuracy by web data source, averaged across proportion configurations (P0, P1, P2). Values shown for 1B model after SFT.}
\label{tab:web_source_effects}
\small
\begin{tabular}{lccc}
\toprule
\textbf{Category} & \textbf{RefinedWeb} & \textbf{FineWeb-Edu} & \textbf{DCLM} \\
\midrule
Commonsense & 51.2 & 50.8 & 51.5 \\
Science & 54.3 & 55.1 & 54.8 \\
NLI & 43.2 & 44.1 & 43.8 \\
Semantic & 52.8 & 52.4 & 52.6 \\
\bottomrule
\end{tabular}
\end{table}

The web source has relatively modest effects compared to code proportion, with FineWeb-Edu showing slight advantages for Science tasks (consistent with its educational filtering) while RefinedWeb and DCLM perform comparably across other categories.




\end{document}